\documentclass{article}
\PassOptionsToPackage{numbers, compress,sort}{natbib}

\usepackage[preprint]{neurips_2022}

\usepackage[utf8]{inputenc} 
\usepackage[T1]{fontenc}    
\usepackage{hyperref}       
\usepackage{url}            
\usepackage{booktabs}       
\usepackage{amsfonts}       
\usepackage{nicefrac}       
\usepackage{microtype}      
\usepackage{xcolor}         
\usepackage{tikz}
\usepackage{pgfplots}
\usepackage{subcaption}
\usepackage{amsmath}
\usepackage{amssymb}
\usepackage{bbding}
\usepackage{mathtools}
\usepackage{amsthm}
\usepackage{multirow}
\usepackage{multicol}
\usepackage[capitalize,noabbrev]{cleveref}
\usepackage{makecell}
\usepackage{hyperref}
\usepackage{url}
\usepackage{times}
\usepackage{soul}
\usepackage{url}
\usepackage{graphicx}
\usepackage{amsmath}
\usepackage{amsthm}
\usepackage{amssymb}
\usepackage{booktabs}
\usepackage{algorithm}
\usepackage{algorithmic}
\usepackage{color}

\usepackage{wrapfig}
\usepackage{natbib}
\usepackage{colortbl}
\definecolor{mygray}{gray}{0.9}

\newcommand{\Cmat}[0]{\ensuremath{{\bf C}} }

\newcommand{\Tmat}[0]{\ensuremath{{\bf T}} }

\newcommand{\Wmat}[0]{\ensuremath{{\bf W}} }

\newcommand{\av}[0]{\ensuremath{\boldsymbol{a}} }
\newcommand{\bv}[0]{\ensuremath{\boldsymbol{b}} }

\newcommand{\sv}[0]{\ensuremath{\boldsymbol{s}} }

\newcommand{\wv}[0]{\ensuremath{\boldsymbol{w}} }

\newcommand{\zv}[0]{\ensuremath{\boldsymbol{z}} }

\newcommand{\Omegamat}[0]{\ensuremath{\boldsymbol{\Omega}}}

\newcommand{\thetav}[0]{\ensuremath{\boldsymbol{\theta}} }

\newcommand{\given}{\,|\,}

\title{Learning to Re-weight Examples with Optimal Transport for Imbalanced Classification}

\author{%
  Dandan Guo\thanks{Use footnote for providing further information
    about author (webpage, alternative address)---\emph{not} for acknowledging
    funding agencies.} \\
  Department of Computer Science\\
  Cranberry-Lemon University\\
  Pittsburgh, PA 15213 \\
  \texttt{hippo@cs.cranberry-lemon.edu} \\
}

\author{Dandan Guo \textsuperscript{1},~ Zhuo Li\textsuperscript{1,2},~ Meixi Zheng\textsuperscript{3},~He Zhao \textsuperscript{4}, ~ Mingyuan Zhou\textsuperscript{5},~
Hongyuan Zha\textsuperscript{1}
\\
\textsuperscript{1}The Chinese University of Hong Kong, Shenzhen~~
\textsuperscript{2} Shenzhen Research Institute of Big Data~~
\\
\textsuperscript{3}Xidian University~~~
\textsuperscript{4}CSIRO's Data61 ~~~
\textsuperscript{5}The University of Texas at Austin\\
\texttt{guodandan@cuhk.edu.cn}\quad \texttt{221019088@link.cuhk.edu.cn}\quad \texttt{meixizheng1110@163.com}\\
\texttt{he.zhao@ieee.org}\quad \texttt{mingyuan.zhou@mccombs.utexas.edu} \quad
\texttt{zhahy@cuhk.edu.cn}
}

\begin{document}
\maketitle

\begin{abstract}
Imbalanced data pose challenges for deep learning based classification models. One of the most widely-used approaches for tackling imbalanced data is re-weighting, where training samples are associated with different weights in the loss function. Most of existing re-weighting approaches treat the example weights as the learnable parameter and optimize the weights on the meta set, entailing expensive bilevel optimization. In this paper, we propose a novel re-weighting method based on optimal transport (OT) from a distributional point of view.
Specifically, we view the training set as an imbalanced distribution over its samples, which is transported by OT to a balanced distribution obtained from the meta set. The weights of the training samples are the probability mass of the imbalanced distribution and learned by minimizing the OT distance between the two distributions. Compared with existing methods, our proposed one disengages the dependence of the weight learning on the concerned classifier at each iteration. Experiments on image, text and point cloud datasets demonstrate that our proposed re-weighting method has excellent performance, achieving state-of-the-art  results in many cases and providing a promising tool for addressing the imbalanced classification issue.

\end{abstract}

\section{Introduction}
Deep neural networks (DNNs) have achieved remarkable success in various applications, which is undoubtedly inseparable from the high-quality large-scale datasets. Usually, the number of samples for each class in these datasets are manually selected resulting in balanced datasets. However, most real-world datasets are imbalanced, such as a few classes (a.k.a. head or majority class) occupy most of the data while most classes (a.k.a. tail or minority class) have a few samples.
A model trained on the imbalanced training
set but without considering such class imbalance
would be significantly dominated by those majority classes, and thus underperform on a balanced test dataset. This can also be known as the long-tailed problem and exists in many domains, such as text classification \cite{hu2019learning,liu2021improving}, object detection \cite{DBLP:journals/pami/OksuzCKA21} and image classification \cite{jamal2020rethinking,li2021metasaug,DBLP:conf/iclr/KangXRYGFK20}.


There are rich research lines to solve the imbalance problem, including re-sampling \cite{DBLP:journals/jair/ChawlaBHK02,DBLP:conf/icic/HanWM05,2003C4,2004The}, class-level or instance-level re-weighting  \cite{2016Learning,lin2017focal,cui2019class,wang2017learning,ren2018learning,jamal2020rethinking,shu2019meta,hu2019learning,liu2021improving,cao2019learning,IB21}, meta-learning \cite{ren2018learning,ren2020balanced,li2021metasaug,jamal2020rethinking,shu2019meta}, two-stage methods \cite{cao2019learning,jamal2020rethinking,li2021metasaug,DBLP:conf/iclr/KangXRYGFK20} and post-hoc correction \cite{menon2020long,peng2022optimal}. Inspired by \cite{liu2021improving}, re-weighting strategies can be roughly grouped into empirical re-weighting and automatic re-weighting. The former aims to design weights manually with the major insight that \textit{the minority class example will be assigned a larger weight value than that of the majority class} \cite{wang2017learning,cui2019class,lin2017focal}. However, manually setting weights can be less adaptive to different datasets \cite{liu2021improving}. The latter aims to assign adaptive weights to the examples through learning mechanisms  \cite{ren2018learning,shu2019meta,jamal2020rethinking,liu2021improving,hu2019learning}. As the representative automatic re-weighting method, L2RW \cite{ren2018learning} optimizes the weight vector as a learnable parameter with an unbiased meta set ($i.e.$, validation set).
Although L2RW and its followers have received widespread attention, most of them may be limited to optimizing the weights by the classification loss on the meta set: The gradient of weights is usually coupled with the to-be-learned classifier at each training iteration. Since classifier is the major concern in imbalanced issue \cite{DBLP:conf/iclr/KangXRYGFK20}, the dependence of weights on classifier at training stage may lead to inaccurate learning of the weights.


This paper develops a novel automatic re-weighting method for imbalanced classification based on optimal transport (OT). As discussed by \citet{jamal2020rethinking}, the major challenge for imbalanced classification is essentially the mismatch between the imbalanced training dataset (seen by a machine learning model) and the balanced test set (used to test the learned model). To this end, we aim to view the learning of the weight vector as the distribution approximation problem. We adopt the two-stage learning manner motivated by \cite{DBLP:conf/iclr/KangXRYGFK20}, where stage 1 and stage 2 focus on learning the feature extractor with the standard cross-entropy loss and the classifier with our proposed method, respectively. Specifically, we represent the imbalanced training set as a discrete empirical distribution $P$ over all samples within it and view the to-be-learned weight vector $\wv$ as its probability measure.
Then we represent the balanced meta set as a discrete empirical distribution $Q$ over all samples within it (in the same space with $P$ ), which has a uniform probability measure for being balanced. Therefore, the learning of a weight vector can be formulated as the process of learning the distribution $P$ to be as close to the balanced distribution $Q$ as possible, a process facilitated by leveraging the OT distance \cite{PeyreC2019OT}. Notably, the cost function plays a paramount role when learning the transport plan for OT, where we use the features and ground-truth labels of samples to design it.
Due to the flexibility of our method, we can also learn an explicit weight net directly from data like {\cite{shu2019meta,jiang2018mentornet}} but with a different structure, optimized by OT loss instead of the classification loss on the meta set. Generally, at each training iteration at stage 2, we minimize the OT loss to learn the weight vector (or weight net) for the current mini-batch, which is further used to re-weight the training  loss for optimizing the model. As we can see, the gradient of weights only relies on the OT loss and thus is independent of the classifier. More importantly, our proposed method is robust to the distribution $Q$. To save the memory consumption, we introduce the prototype-oriented OT loss by building a new distribution $Q$ based on prototypes instead of samples (one prototype for each class). More importantly, our proposed method can achieve a reasonably good performance even if we randomly select a mini-batch from all prototypes to build $Q$, making our method applicable to datasets with a large number of classes.




We summarize our main contributions as follows: (1) We formulate the learning of weight vector or weight net as the distribution approximation problem by minimizing the statistical distance between to-be-learned distribution over samples from imbalanced training set and another balanced distribution over samples from the meta set. (2) We leverage the OT distance between the distributions to guide the learning of weight vector or weight net. (3) We apply our method to imbalanced classification tasks including image, text and point cloud. Experiments demonstrate that introducing the OT loss to learn the example weights can produce effective and efficient classification performance.

\section{Related Work}

\textbf{Empirical Re-weighting} A classic empirical re-weighting scheme is to provide the examples of each class with the same weight, such as inverse class frequency \cite{2016Learning,wang2017learning}. It has been further improved by the class-balanced loss \cite{cui2019class}, which calculates the effective number of examples as class frequency. Focal Loss \cite{lin2017focal} uses the predicted probability to calculate higher weights for the hard examples and dynamically adjust the weights. LDAM-DRW \cite{cao2019learning} designs a label-distribution-aware loss function and adopts a deferred class-level re-weighting method  ($i.e.$, inverse class frequency).

\textbf{Automatic Re-weighting}
The automatic re-weighting methods learn the weights with learning mechanisms. L2RW \cite{ren2018learning} adopts a meta-learning manner to learn the example weights, which are optimized by the classification loss on the balanced meta set. \citet{hu2019learning} further improve L2RW by iteratively optimizing weights instead of re-estimation at each iteration. Meta-weight-net \cite{shu2019meta} aims to learn an explicit weight net directly from data and optimize it by a meta-learning manner. Meta-class-weight \cite{jamal2020rethinking} defines the weight for each example as the combination of class-level weight (estimated by \citet{cui2019class}) and instance-level weight, optimized with a meta-learning approach similar to L2RW. {Influence-balanced loss (IB) is proposed to \cite{IB21} re-weight samples by the magnitude of the gradient.} Recently, \citet{liu2021improving} propose to update the weights and model under a constraint. Our method belongs to automatic re-weighting group, and the idea of building an explicit weight net is similar to \citet{shu2019meta}. However, the major difference is that we bypass  the classification loss on the meta set and use OT to learn the weights from the view of distribution approximation, disengaging the dependence of the weight learning on the concerned classifier at each iteration.

\textbf{Meta Learning and Two-stage Learning}
Recently, researchers have proposed to tackle the imbalance issue with meta-learning, which can be applied to {build a Balanced Meta-Softmax (BALMS)\cite{ren2020balanced}}, learn weights \cite{ren2018learning,shu2019meta,jamal2020rethinking} or transformed semantic directions for augmenting the minority classes in MetaSAug \cite{li2021metasaug}. Two-stage methods, where the first stage and second stage focus on representation learning and classifier learning, respectively, have been proved effective for solving the imbalanced issue \cite{shu2019meta,li2021metasaug,liu2019large,DBLP:conf/iclr/KangXRYGFK20}. BBN \citep{zhou2020bbn} unifies two stages with a specific cumulative learning strategy.

\textbf{Optimal Transport}
{
Recently, OT has been used to solve the regression problem under the covariate shift \cite{reygner2022reweighting}, unsupervised domain adaption \cite{rakotomamonjy2022optimal,turrisi2020multi}, including sample-level, class-level or domain-level weight vector. Although they also adopt the re-weighting strategy and OT distance, they are distinct form ours in terms of task and technical detail. Also, the dynamic importance weighting which adopts MMD to re-weight samples for label-noise and class-prior-shift tasks \cite{fang2020rethinking} is also different from ours, where we provide a more flexible way for learning the weights of samples and disengage the dependence of the weight learning on the concerned classifier at each iteration.}
To the best of our knowledge, the works that solve imbalanced classification problem with OT are still very limited. An oversampling method via OT (OTOS) \cite{DBLP:conf/aaai/YanTXCNMW19} aims to make synthetic samples follow a similar distribution to that of minority class samples. However, ours is a novel re-weighting method based on OT, without augmenting samples. Another recent work is Optimal Transport via Linear Mapping (OTLM) \cite{peng2022optimal}, which performs the post-hoc correction from the OT perspective and proposes a linear mapping to replace the original exact cost matrix in OT problem. Different from OTLM that belongs to the post-hoc correction group and aims to learn refined prediction matrix, ours falls into the training-aware group and aims to re-weight the training classification loss.

\section{Background}




\textbf{Imbalanced Classification} Consider a training set $\mathcal{D}_{\text {train }}\!=\!\left\{\left(x_{i}, y_{i}\right)\right\}_{i=1}^{N}$, where $(x,y)$ is the input and target pair, $x_i$ the $i$-th sample, $y_i \in(0,1)^{K}$ the one-hot associated label vector over $K$ classes, and $N$ the number of the entire training data. Besides, consider a small balanced meta set $\mathcal{D}_{\text {meta }}\!=\!\left\{\left(x_{j}, y_{j}\right)\right\}_{j=1}^{M}$, where $M$ is the amount of total samples and $M \!\ll\! N$. Denote the model parameterized with $\thetav$ as $f(x, \thetav)$, where $\thetav$ is usually optimized by empirical risk minimization over the training set, $i.e.$, $\thetav^{*}=\arg \min _{\thetav} \frac{1}{N} \sum_{i=1}^{N} \ell\left(y_{i}, f\left(x_{i} ; \thetav \right)\right)$. For notational convenience, we denote $l_{i}^{\text{train}}(\thetav)=\ell\left(y_{i}, f\left(x_{i} ; \thetav \right)\right)$ to represent the training loss function of pair $(x_{i}, y_i)$. However, the model trained by this method will prefer the majority class if the training dataset is imbalanced.



\textbf{Learning to Re-Weight Examples} To solve the imbalanced issue, a kind of re-weighting methods is to treat the weights as the learnable parameter and learn a fair model to the minority and the majority classes by optimizing the weighted training loss. At each training iteration, the model is updated by
\begin{equation}\label{optimize_theta}
\thetav^{*}(\wv)=\arg \min _{\thetav}   \sum\nolimits_{i=1}^{N}  w_{i} l_{i}^{\text{train}}(\theta),
\end{equation}
where $\wv\!=\!\left(w_{1}, \ldots, w_{N}\right)^{T}$ is the  weight vector (usually with a simplex constraint) of all training examples.
Then the optimal $\wv$ is obtained by making the model parameter $\thetav^*(\wv)$ from Eq. \eqref{optimize_theta} minimize the classification loss on a balanced meta set, formulated as
\begin{equation}\label{optimize_w}
\wv^{*}=\arg \min _{\wv} \frac{1}{M}  \sum\nolimits_{j=1}^{M} l_{j}^{\text{meta}}\left(\thetav^{*}(\wv)\right),
\end{equation}
where $l_{j}^{\text{meta}}$ is the loss function of pair $(x_{j}, y_j)$ from meta set and the updated $\wv^{*}$ is used to ameliorate the model. Generally, model $\thetav$ consists of two key components, feature extractor and classifier, where the classifier has been proved to be the major concerning part in imbalanced issue \cite{DBLP:conf/iclr/KangXRYGFK20}. However, the gradient of weights in Eq. \eqref{optimize_w}  always depends on the to-be-concerned classifier at each training iteration, which may result in inaccurate learning of the weights. Most automatic re-weighting methods learn the weight vectors or weight-related parameters ($e.g.$, weight net) following this line; see more details from the previous
works \cite{ren2018learning,shu2019meta,jamal2020rethinking}.

\textbf{Optimal Transport Theory}  OT has been widely used to calculate the cost of transporting one probability measure to another. Among the rich theory of OT, this work presents a brief introduction to OT for discrete distributions; see \citet{PeyreC2019OT} for more details. Consider $p=\sum_{i=1}^{{n}} a_{i} \delta_{x_{i}}$ and $q=\sum_{j=1}^{m} b_{j} \delta_{y_{j}}$ as two probability distributions, where ${x_{i}}$ and ${y_{j}}$ live in the arbitrary same space and $\delta$ is the Dirac function. Then, we can denote $\av \in \Delta^{n}$ and $\bv \in \Delta^{m}$ as the probability simplex of~$\mathbb{R}^{n}$ and ~$\mathbb{R}^{m}$, respectively. The OT distance between $p$ and $q$ can be expressed as:
\begin{equation}\label{OT}
\text{OT}(p, q)=\min _{\mathbf{T} \in \Pi(p, q)}\langle\mathbf{T}, \mathbf{C}\rangle,
\end{equation}
where $\langle\cdot, \cdot\rangle$ is the Frobenius dot-product and $\mathbf{C} \in \mathbb{R}_{\geq 0}^{n \times m}$ is the transport cost matrix constructed by $C_{ij}\!=\!C(x_i,y_j)$. The doubly stochastic transport probability matrix $\mathbf{T}\in \mathbb{R}_{>0}^{n \times m}$, which satisfies $\Pi(p, q):\!=\!\{\mathbf{T} \given \sum_{i=1}^{n} T_{ij}\!=\!b_{j},\sum_{j=1}^{m} T_{ij}\!=\!a_{i}\}$, is learned by minimizing  $\text{OT}(p, q)$.
Directly optimizing Eq. \eqref{OT} often comes at the cost of heavy computational demands, and OT with entropic regularization is introduced to allow the optimization at small computational cost in sufficient smoothness \cite{cuturi2013sinkhorn}.

\section{Re-weighting Method with Optimal Transport}
This work views a training set as a to-be-learned distribution, whose probability measure is set as learnable weight vector $\wv$. We use OT distance to optimize $\wv$ for re-weighting the training loss.

\subsection{Main Objective}
Given the imbalanced training set $\mathcal{D}_{\text {train }}$, we can represent it as an empirical distribution over $N$ pairs, where each pair $(x_{i},y_i)^{\text{train}}$ has the sample probability $w_i$ ($i.e.$, the weight), defined as:
\begin{equation}\label{P_distribution}
P(\wv)={\sum\nolimits_{i=1}^{N} w_{i} \delta_{(x_{i},y_{i})^{\text{train}}}},
\end{equation}
where $(x_{i},y_{i})^{\text{train}}$ is the $i$-th pair from the training set and the learnable weight vector $\wv$ of all training examples means probability simplex of~$\mathbb{R}^{N}$.
Since the meta set $\mathcal{D}_{\text {meta }}$ is balanced for all classes and closely related with the training set, it is reasonable to assume that meta set has already achieved the balanced data distribution that the training set aims to approximate. For meta set, we thus can sample each pair from it with equal probability and present it with an empirical distribution $Q$:
\begin{equation}\label{Q_distribution}
Q= {\sum\nolimits_{j=1}^{M}\frac{1}{M} \delta_{(x_j,y_j)^{\text{meta}}}},
\end{equation}
where $(x_j,y_j)^{\text{meta}}$ is the $j$-th pair from the meta set. To learn $\wv$, different from most automatic re-weighting methods, which minimize the classification loss on the meta set, we aim to enforce the to-be-learned distribution $P(w)$ to stay close to the balanced distribution $Q$. Here, we explore the re-weighting method by adopting the OT distance between $P(w)$ and $Q$:

\begin{equation}\label{OT_our}
\min_{\wv}\text{OT}(P(\wv), Q)\stackrel{\text { def. }}{=} \min_{\wv}\min_{\mathbf{T} \in \Pi(P(\wv), Q)} \langle \Tmat, \Cmat \rangle,
\end{equation}

where cost matrix $\Cmat \!\in\! \mathbb{R}_{\geq 0}^{N \times M}$ is  described below and transport probability matrix $\Tmat \!\in \!\mathbb{R}_{>0}^{N \times M}$  should satisfy $\Pi(P(\wv), Q): =\{\mathbf{T} \given \sum_{i=1}^{N} T_{ij}\!=\!1/M,\sum_{j=1}^{M} T_{ij}\!=\!w_{i}\}$.


\subsection{Cost Function}
For notation convenience, we reformulate the model as $f(x, \thetav) =f_{2}(f_{1}(x;\thetav_{1});\thetav_{2})$, where $f_{1}$ parameterized with $\thetav_{1}$ denotes the representation learning part before the classifier, and $f_{2}$ parameterized with $\thetav_{2}$ denotes the classifier.
Intuitively, the cost $C_{ij}$ measures the distance between pair $i$ in training set and pair $j$ in meta set, which can be flexibly defined in different ways. We explore a few conceptually intuitive options of $C_{ij}$, although other reasonable choices can also be used.

\textbf{Label-aware Cost}
As the first option, we can define $C_{ij}$ with the ground-truth labels of two samples:
\begin{equation}\label{OT_sink2}
C_{ij}=\operatorname{d}^{\text{Lab}}(y_{i}^{\text{train}},y_{j}^{\text{meta}}),
\end{equation}
where $\operatorname{d}^{\text{Lab}}(\cdot, \cdot)$ also denotes a distance measure, and $y_{i}^{\text{train}}, y_{j}^{\text{meta}}$ are the ground-truth label vectors of the two samples, respectively.
Intuitively, if we use the euclidean distance, then $\Cmat$ is a $0\!-\!1$ matrix (we can transfer the non-zero constant to $1$),
$i.e.$, $C_{ij}\!=\!0$ if $x_{i}^{\text{train}}$ and $x_{j}^{\text{meta}}$ are from the same class, and $C_{ij}\!=\!1$ otherwise. Now the OT loss is  influenced by neither feature extractor $\thetav_1$ nor classifier $\thetav_2$.

\textbf{Feature-aware Cost}
Besides, we can define $C_{ij}$ purely based on the features of samples:
\begin{equation}\label{OT_sink1}
C_{ij}=\operatorname{d}^{\text{Fea}}(\zv_{i}^{\text{train}},\zv_{j}^{\text{meta}}),
\end{equation}
where $\zv_{i}^{\text{train}}\!=\!f_{1}(x_{i}^{\text{train}};\thetav_{1}) \in \mathbb{R}^{E}$ and $\zv_{j}^{\text{meta}}\!=\!f_{1}(x_{j}^{\text{meta}};\thetav_{1})\in \mathbb{R}^{E}$ denote the $E$-dimensional representation of $x_{i}^{\text{train}}$ and $x_{j}^{\text{meta}}$, respectively.
$\operatorname{d}^{\text{Fea}}(\cdot, \cdot)$ denotes any commonly used distance measure and we empirically find the cosine distance is a good choice.
It is easy to see that if $x_{i}^{\text{train}}$ and $x_{j}^{\text{meta}}$'s features are close, their cost is small.
Here the OT loss is influenced by the feature extractor $\thetav_1$.

\textbf{Combined Cost}
Finally, we can use both features and labels to define $C_{ij}$,
denoted as
\begin{equation}\label{OT_sink_combine}
C_{ij}=\operatorname{d}^{\text{Fea}}(\zv_{i}^{\text{train}},\zv_{j}^{\text{meta}})+\operatorname{d}^{\text{Lab}}(y_{i}^{\text{train}},y_{j}^{\text{meta}}).
\end{equation}
Intuitively, $C_{ij}$ will be small if two samples have the same label and similar features. Empirically, we find that using the $\operatorname{d}^{\text{Fea}}\!=\!1\!-\!\operatorname{cosine}(\cdot, \cdot)$ and  euclidean distance for $\operatorname{d}^{\text{Lab}}$ gives better performance. Interestingly, given the feature-aware cost \eqref{OT_sink1} or label-aware cost \eqref{OT_sink2}, the learned weight vector can be interpreted as the instance-level or class-level re-weighting method, respectively. The weight vector learned from the combined cost can be interpreted as the combination of class-level and instance-level weights, although no specialized design for two-component weights like previous \cite{jamal2020rethinking}; see Fig. \ref{fig:cost_weight}.

\subsection{Learn the Weight Vector}
Given the defined cost function, we adopt the entropy regularized OT loss \cite{cuturi2013sinkhorn} to learn the weight vector. We thus rewrite \eqref{OT_our} as the following optimization problem:
\begin{equation}\label{OT_loss_final}
\vspace{-1mm}
\min _{\wv} L_{\text{OT}}=\left\langle \Cmat, \Tmat_{\lambda}^{*}(\wv)\right\rangle \text {, subject to } \Tmat_{\lambda}^{*}(\wv)=\underset{\Tmat \in \Pi\left(P(\wv), Q\right)}{\arg \min }\langle \Tmat, \Cmat\rangle-\lambda H(\Tmat),
\end{equation}
where $\lambda \textgreater 0$ is a hyper-parameter for the entropic constraint $H(\Tmat)\!=\!-\sum\nolimits_{ij} T_{ij}\textrm{ln}T_{ij}$. Note that \eqref{OT_loss_final} provides us a new perspective to interpret
the relationship between $\wv$ and $\Tmat$, where $\wv$ is the parameter of the leader problem and $\Tmat$ is the parameter of the follower problem, which is of the lower priority. Accordingly, when we minimize \eqref{OT_loss_final}  with respect to $\wv$ using gradient descent, we should differentiate through $\Tmat$.
Below we investigate the following two ways to optimize the weight vector.

\textbf{Optimizing $\wv$ directly}
Specifically, at each training iteration, we define $P(\wv)$ with current $\wv$, use the Sinkhorn algorithm \cite{cuturi2013sinkhorn} to compute OT loss, then optimize $\wv$ by $\wv^{*}=\arg \min _{\wv}L_{\text{OT}}$.

\textbf{Amortizing the learning of $\wv$} We also provide an alternative method by constructing an explicit weight net to output the example weights, whose structure can be designed flexibly. For example, we can build the following weight net and take the sample features as input:
\begin{align}\label{atten}
\wv =\operatorname{softmax}\left(\sv\right), s^{i} \!=\! \wv_{att} \tanh \left(\Wmat_{vz} \zv_{i}^{\text{train}}\! \right) ,
\end{align}
where $s^{i}$ is the $i$-th element of $\sv \in \mathbb{R}^{N} $,  $ \wv_{att} \in \mathbb{R}^{1 \times A} $ and $ \Wmat_{vz} \in \mathbb{R}^{A \times E} $ are the learned parameters (we omit the bias for convenience), denoted as $\Omegamat=\{\wv_{att},\Wmat_{vz}\}$.
Denote $S(\zv;\Omegamat)$ as the weight net parameterized by $\Omegamat$, which can be optimized by  $\Omegamat^{*}=\arg \min _{\Omegamat}L_{\text{OT}} $.





\section{Overall Algorithm and Implementations}
To integrate our proposed method with deep learning frameworks, we adopt a stochastic setting, $i.e.$, a mini-batch setting at each iteration. Following \cite{jamal2020rethinking,li2021metasaug}, we adopt two-stage learning, where stage 1 trains the model $f(\thetav)$ by the standard cross-entropy loss on the imbalanced training set and stage 2 aims to learn the weight vector $\wv$ and meanwhile continue to update the model $f(\thetav)$.
Generally, at stage 2, calculating the optimal $\thetav$ and $\wv$ requires two nested loops of optimization, which is cost-expensive. Motivated by \citet{hu2019learning}, we optimize $\thetav$ and $\wv$ alternatively, corresponding to \eqref{optimize_theta} and \eqref{OT_loss_final} respectively, where $\wv$ is maintained and updated throughout the training, so that  re-estimation from scratch can be avoided in each iteration. The implementation process of our proposed method with $\wv$ optimized directly is shown in Algorithm \ref{Algorithm1}, where
the key steps are highlighted in Step (a), (b), and (c).
Specifically,
at each training iteration $t$, in Step (a), we have  $\hat{\thetav}^{(t+1)}(\wv^{t})=\{\hat{\thetav}_1^{(t+1)}(\wv^{t}),\hat{\thetav}_2^{(t+1)}(\wv^{t})\}$ and $\alpha$ is the step size for $\thetav$; in Step (b), as the cost function based on features is related with $\hat{\thetav}_1^{(t+1)}(\wv^{t})$, the OT loss relies on $\hat{\thetav}_1^{(t+1)}(\wv^{t})$, and $\beta$ is the step size for $\wv$; in Step (c), we ameliorate model parameters $\thetav^{(t+1)}$. We defer the learning of $\thetav$ and $\Omegamat$ for the amortized learning of $\wv$.



\textbf{Discussion} From Step (b), we find the gradient of $\wv$ is unrelated to classifier $\thetav_2$ regardless of which cost function we choose. If we use the label-aware cost or freeze the feature extractor parameterized by $\thetav_1$, which is trained in the first stage, the OT loss in Step (b) can be further reduced as $L_\text{OT}\left(\wv^{t}\right)$, where we only need Steps (b)-(c) at each iteration. This is different from most of automatic re-weighting methods, where the gradient of $\wv$ is always related with the to-be-learned model $\{\thetav_1, \thetav_2\}$ or classifier $\thetav_2$ (when freezing $\thetav_1$) for minimizing the classification loss on meta set.

\textbf{Prototype-oriented OT loss (POT)}
Recall that we represent a balanced meta set with $M$ samples as distribution $Q$ in \eqref{Q_distribution}, where $M/K$ is the number of data in each class and usually larger than 1. Computing the OT loss requires to learn a $B \times M$-dimensional transport matrix at each iteration.
To improve the efficiency of algorithm, we average all samples from each class in the meta set to achieve its prototype and propose a new $Q$ distribution over $K$ prototypes:
\begin{equation}\label{new_Q_distribution}
Q= {\sum\nolimits_{k=1}^{K}\frac{1}{K} \delta_{(\hat{x}_{k},y_k)^{\text{meta}}}},\quad \hat{x}_{k}=\frac{K}{M}\sum\nolimits_{j=1}^{M/K} x_{kj}^{\text{meta}}.
\end{equation}
where POT loss only needs a $B \times K$-dimensional transport matrix. Due to the robustness of our method to $Q$, when dealing with a large number of classes, we can randomly sample a mini-batch from $K$ prototypes at each iteration to build $Q$.

\begin{algorithm}[!t]
\caption{\footnotesize{{Workflow about our re-weighting method for optimizing $\thetav$ and $\wv$.}}}
\begin{algorithmic}
 \STATE \textbf{Require:} Datasets $\mathcal{D}_{\text {train }}$, $\mathcal{D}_{\text {meta }}$, initial model parameter $\thetav$ and weight vector, hyper-parameters $\{\alpha,\beta,\lambda\}$
\FOR{$t=1,2,...,t_1$}
  \STATE Sample a mini-batch $B$ from the training set $\mathcal{D}_{\text {train }}$;
  \STATE Update $\thetav^{(t+1)} \leftarrow \thetav^{(t)}-\alpha \nabla_{\thetav} \mathcal{L}_{B}$ where $\mathcal{L}_{B}=\frac{1}{|B|} \sum_{i \in B} \ell\left(y_{i}, f\left(x_{i} ; \thetav^{(t)} \right)\right) $;
\ENDFOR
\FOR{$t=t_1+1,...,t_1+t_2$}
  \STATE Sample a mini-batch $B$ from the training set $\mathcal{D}_{\text {train }}$;
  \STATE \textbf{Step (a):} Update $\hat{\thetav}^{(t+1)}(\wv^{(t)}) \gets \thetav^{(t)}-\alpha \nabla_{\thetav} \mathcal{L}_{B}$ where $\mathcal{L}_{B}=\frac{1}{|B|} \sum_{i \in B} w_i^{(t)} \ell\left(y_{i}, f\left(x_{i} ; \thetav^{(t)} \right)\right)$
      \STATE Use $\mathcal{D}_{\text {meta }}$ to build $Q$ in \eqref{new_Q_distribution} and $B$ with $\wv^{t}$ to build  $P(\wv^{t})$ \eqref{P_distribution};
  \STATE \textbf{Step (b):} Compute $L_{\text{OT}} \left(\hat{\thetav}_1^{(t+1)}(\wv^{t}),\wv^{(t)}\right)$ with cost \eqref{OT_sink_combine}; Optimize $\wv^{(t+1)} \gets \wv^{(t)}-\beta \nabla_{\wv} L_{\text{OT}} \left(\hat{\thetav}_1^{(t+1)}(\wv^{t}),\wv^{(t)}\right)$
  \STATE \textbf{Step (c):} Update ${\thetav}^{(t+1)}\gets \thetav^{(t)}-\alpha \nabla_{\thetav} \mathcal{L}_{B}$ where  $\mathcal{L}_{B}=\frac{1}{|B|} \sum_{i \in B} w_i^{(t+1)} \ell\left(y_{i}, f\left(x_{i} ; \thetav^{(t)} \right)\right) $
  \ENDFOR
\end{algorithmic}\label{Algorithm1}
\end{algorithm}

\section{Experiments}
\label{Experiments}

We conduct extensive experiments to validate the effectiveness of our proposed method on text, image, and point cloud imbalanced classification tasks. Notably, different from the imbalanced image and point cloud classification, we find that optimizing the weight net is better than optimizing the weight vector directly in the text classification. Therefore, we optimize the weight vector for the image and point cloud cases and build a weight net for text case. Unless specified otherwise, we adopt the combined cost and set the hyper-parameter for the entropic constraint as $\lambda=0.1$ and the maximum iteration number in the Sinkhorn algorithm as $200$. We define the imbalance factor (\textrm{IF}) of a dataset as the data point amount ratio between the largest and smallest classes.




\subsection{Experiments on Imbalanced Image Classification}
\label{Experiments on Image Imbalanced Classification}
\textbf{Datasets and Baselines}
We evaluate our method on CIFAR-LT-10, CIFAR-LT-100, ImageNet-LT and Places-LT. We create \textbf{\textit{CIFAR-LT-10}} (\textbf{\textit{CIFAR-LT-100}}) from CIFAR-10 (CIFAR-100)\cite{krizhevsky2009learning} by downsampling samples per class
 with $\textrm{IF}\!\in\!\{200, 100, 50, 20\}$ \citep{li2021metasaug,cui2019class}. \textbf{\textit{ImageNet-LT}} is built from the classic ImageNet  with 1000 classes\cite{deng2009imagenet} and $\textrm{IF}\!=\!1280/5$  \cite{li2021metasaug,liu2019large}.
\textbf{\textit{Places-LT}} is created from  Places-2 \cite{zhou2017places}  with 365 classes and $\textrm{IF}\!=\!4980/5$  \cite{jamal2020rethinking,liu2019large}. We randomly select 10 training images per class as meta set \citep{li2021metasaug}; see more details in Appendix \ref{sec:details_image}. We consider the following baselines: (1) \textbf{Cross-entropy (CE)}, the model trained on the imbalanced training set with CE loss. (2) \textbf{Empirical re-weighting methods}, like Focal loss \citep{lin2017focal}, Class-balanced (CB) loss \citep{cui2019class} and LDAM-DRW \cite{cao2019learning}.
(3) \textbf{Automatic re-weighting methods}, including L2RW \cite{ren2018learning}, {IB \cite{IB21}}, Meta-Weight-Net \cite{shu2019meta} and Meta-class-weight \cite{jamal2020rethinking}.
(4) \textbf{Meta-learning methods}, including MetaSAug \cite{li2021metasaug} and above methods of \cite{ren2018learning,shu2019meta,jamal2020rethinking,ren2020balanced}.
(5) \textbf{Two-stage methods}, such as OLTR \cite{liu2019large}, cRT \cite{DBLP:conf/iclr/KangXRYGFK20}, LWS \cite{DBLP:conf/iclr/KangXRYGFK20}, BBN \cite{zhou2020bbn} and methods of \cite{jamal2020rethinking,li2021metasaug}.







\textbf{Experimental details and results on CIFAR-LT}
For a fair comparison, we use ResNet-32 \cite{he2016deep} as the backbone on CIFAR-LT-10 and CIFAR-LT-100.
Following \citet{li2021metasaug}, at stage 1, we use $200$ epochs, set the learning rate $\alpha$ of $\thetav$ as 0.1, which is decayed by $1e^{-2}$ at the 160th and 180th epochs. At stage 2, we use $40$ epochs, set $\alpha$ as $2e^{-5}$ and learning rate $\beta$ of weights as $1e^{-3}$. We use the SGD optimizer with momentum 0.9, weight decay $5e^{-4}$ and set the batch size as $16$. We list the recognition results of different methods on CIFAR-LT-10 and CIFAR-LT-100 with different imbalance factors in Table \ref{cifar10}.We report the average result of 5 random experiments without standard deviation which is of small scale(e.g., 1e-2). We can see that our re-weighting method outperforms CE training by a large margin and performs better than the empirical or automatic re-weighting methods. Remarkably, our proposed method outperforms competing MetaSAug that conducts a meta semantic augmentation approach to learn appropriate class-wise covariance matrices when $\text{IF}$ is 200, 100 and 50. Importantly, as the training data becomes more imbalanced, our method is more advantageous. Even though our proposed method is inferior to MetaSAug when the dataset is less imbalanced ($\text{IF}\!=\!20$), it can still achieve competing results and surpasses related re-weighting methods. This suggests that our proposed method can be used to enhance the imbalanced classification, without the requirement of designing complicated models or augmenting samples on purpose.


To more comprehensively understand our method, we provide a series of ablation studies on CIFAR-LT-100 with $\text{IF}\!=\!200$ in Table \ref{ablation}. Firstly, to explore the impact of cost function, we use different cost functions for the OT loss. We can see that the combined cost performs better than label-aware cost and feature-aware cost, confirming the validity of combining features and labels to define cost. Besides, using either label-aware or feature-aware cost can still achieve acceptable performance, indicating the usefulness of OT loss in the imbalanced issue. Secondly, to explore the robustness of the meta distribution $Q$, we adopt three ways to build $Q$: (1) using prototypes defined in Eq. \eqref{new_Q_distribution} ($K$ samples) ; (2) using all samples defined in Eq. \eqref{Q_distribution} ($10*K$ samples) ; (3) randomly sampling one point from each class ($K$ samples) in meta set. We find that prototype-based meta performs best, and the performance with random-sample meta or whole meta is still competitive, which demonstrates the robustness of our proposed method to the distribution $Q$ and the benefit of using the prototypes to build $Q$. Third, we compare two ways for learning  $\wv$ in each iteration, where one is re-estimating $\wv$ from scratch and another one is maintaining and updating $\wv$ throughout the training ($i.e.$, iteratively optimizing weights). We find that   iteratively optimizing performs better.

\begin{table}[!ht]
\centering
\vspace{-4mm}
\caption{\small{Test top-1 errors (\%) of ResNet-32 on CIFAR-LT-10 and CIFAR-LT-100 under different settings. }}
\resizebox{0.99\textwidth}{!}{
\begin{tabular}{c|cccc|cccccc}
\toprule[1pt]
\multicolumn{1}{c|}{\textbf{Datasets}}&\multicolumn{4}{c|}{\textbf{CIFAR-LT-10}} &\multicolumn{4}{c}{\textbf{CIFAR-LT-100}} \\\hline
\textbf{Imbalance Factor} & 200 & 100 & 50 & 20&200&100&50&20 \\ \hline
{CE loss(results from \cite{li2021metasaug})}&  34.13 & 29.86 & 25.06 & 17.56 & 65.30& 61.54 &55.98 &48.94\\ \hline
Focal loss \cite{lin2017focal} (results from \cite{jamal2020rethinking}) &34.71& 29.62 &23.29 &17.24& 64.38& 61.59 &55.68& 48.05  \\
{CB, CE loss} \cite{cui2019class} (results from \cite{li2021metasaug}) & 31.23  &27.32  &21.87 & 15.44 &64.44& 61.23& 55.21 &48.06\\
CB, Focal loss  \cite{cui2019class} (results from \cite{jamal2020rethinking}) &31.85 &25.43 &20.78& 16.22& 63.77 &60.40& 54.79 &47.41\\
LDAM loss \cite{cao2019learning} (results from \cite{li2021metasaug}) &33.25& 26.45& 21.17 &16.11& 63.47& 59.40 &53.84& 48.41\\
LDAM-DRW \cite{cao2019learning} (results from \cite{li2021metasaug}) &25.26 &21.88& 18.73& 15.10 &61.55& 57.11 &52.03 &47.01\\\hline
L2RW \cite{ren2018learning} (results from \cite{shu2019meta})&33.49&25.84& 21.07&16.90&66.62&  59.77&55.56&48.36\\
$\text{Meta-weight net}$ \cite{shu2019meta} &32.80 &26.43 &20.90 &15.55 &63.38 &58.39& 54.34 &46.96\\
Meta-class-weight with CE loss \cite{jamal2020rethinking} &29.34& 23.59& 19.49& 13.54& 60.69 & 56.65&  51.47 & 44.38\\
Meta-class-weight with focal loss \cite{jamal2020rethinking} &25.57 &21.10 &17.12 &13.90& 60.66 &55.30& 49.92 &44.27\\
Meta-class-weight with LDAM loss \cite{jamal2020rethinking} & 22.77& 20.00 &17.77& 15.63 &60.47 &55.92& 50.84& 47.62\\
MetaSAug with CE loss \cite{li2021metasaug}  &23.11 &19.46& 15.97 &12.36& 60.06 &53.13& 48.10 &\textbf{42.15}\\
{IB \cite{IB21}}&{27.85}&{23.47}&{18.34}&{14.59}&{60.34}&{54.61}&{51.07}&{46.43}
\\
{IB+CB \cite{IB21}}&{30.04}&{24.03}&{17.91}&{14.73}&{60.31}&{54.73}&{51.20}&{46.58}
\\
{IB + Focal loss \cite{IB21}}&{25.88}&{22.03}&{17.62}&{14.32}&{59.61}&{55.04}&{51.08}&{45.47}
\\\hline
MetaSAug with focal loss \cite{li2021metasaug}  &22.73 &19.36 &15.96& 12.84 &59.78& 54.11 &48.38 &42.41\\
MetaSAug with LDAM loss \cite{li2021metasaug} &22.65 &19.34 &15.66 &\textbf{11.90} &56.91 &51.99 &47.73 &42.47\\\hline
BBN \cite{zhou2020bbn} &-& 20.18& 17.82& - &-& 57.44& 52.98 &-\\\hline
 \rowcolor{mygray}
\textbf{Our method (Weight Vector)}&\textbf{21.54} & \textbf{18.13} &\textbf{ 15.54}& 12.50&\textbf{54.97}&\textbf{51.46}&\textbf{47.50}&42.85\\
\bottomrule
\end{tabular}}\label{cifar10}
 \vskip -1em
\end{table}

 \begin{figure}[h]
 \centering
\includegraphics[height=3.6cm,width=15cm]{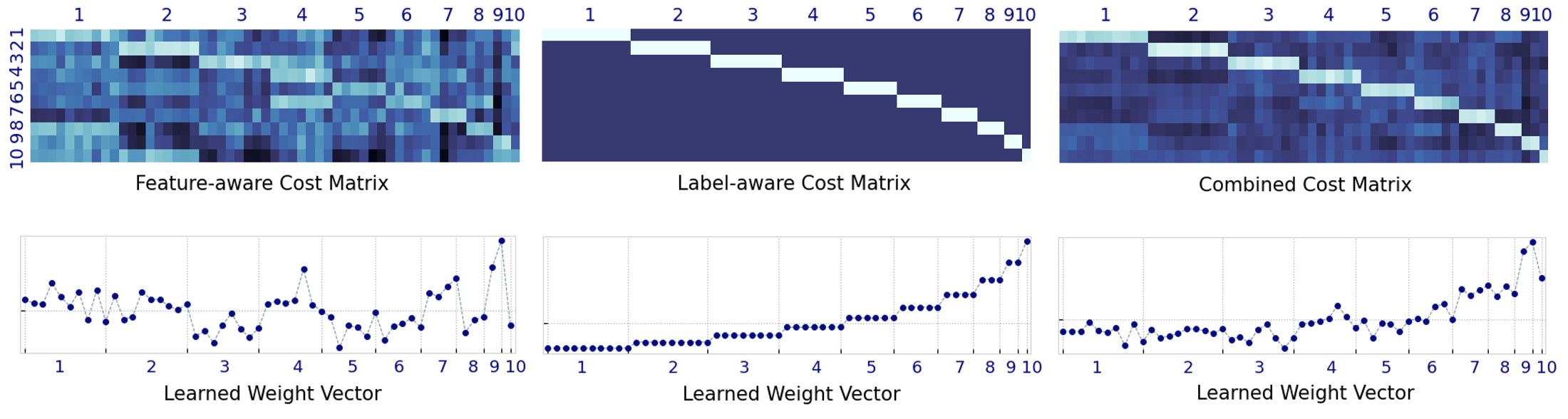}
 \caption{\small{Learned weight vectors (bottom) given different cost functions (top) on CIFAR-LT-10, where x-axis denotes the $55$ samples from the current mini-batch and where we only mark their labels  for clarity.}}\label{fig:cost_weight}
\end{figure}
Since cost function is essential in optimizing the OT loss, we are interested in examining the learned weight vectors given by different cost functions. Here, we use CIFAR-LT-10, randomly choose $\{10,9,...,1\}$ training samples from class $\{1,2,...,10\}$ and obtain $55$ samples, which are used to build distribution $P$. Besides, the $10$ prototypes from meta set are used to build the distribution $Q$. Given the different cost functions, we show the learned weight vectors of $55$ training samples in Fig. \ref{fig:cost_weight}, which have very different properties. Specifically, the label-aware cost and feature-aware cost lead to class-level weights and sample-level weights, respectively. It is reasonable that label-aware cost only decides whether the two samples (from the meta set and training set) belong to the same class, resulting in class-level measure. However, feature-aware cost measures the distance between samples from the sample-level, where each sample has its own feature. More interestingly, the learned weights with the combined cost own the characteristics of class-level and sample-level weights simultaneously, where example weights of different classes are far away and example weights of the same class are close. Coincidentally, using the combined cost to define the OT loss can reach the same goal of \cite{jamal2020rethinking}, which explicitly considers class-level and sample-level weight. Besides, we find that the learned example weights of the minority class are usually more prominent than those of the majority classes.

\begin{figure}[h]
 \centering
\includegraphics[height=4.5cm]{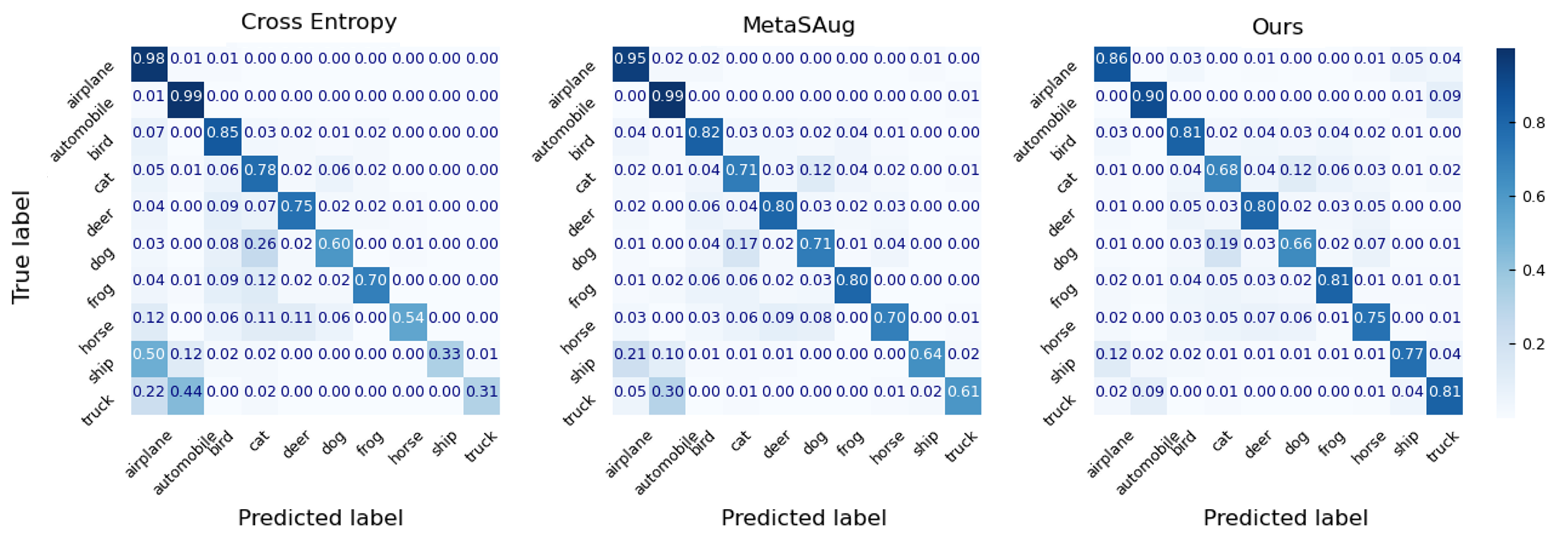}
 \caption{\small{Confusion matrices of the cross-entropy training, MetaSAug and ours on CIFAR-LT-10 with the imbalance factor $200$. We rank classes by the frequency, $i.e.$, frequent (left) and rare (right).}}\label{fig:Confusion}
\end{figure}

\begin{table}[htbp!]
\begin{minipage}{0.33\linewidth}
\caption{\small{Ablation study on CIFAR-LT-100 with $\text{IF}\!=\!200$, where $\wv$ is maintained and updated throughout the training except the last row.}}\label{ablation}
\centering
\resizebox{1\textwidth}{!}
{
\begin{tabular}{c|cc}
\toprule[1pt]
\textbf{Method} &  \textbf{Top-1 errors}\\ \hline
Label +Prototype & 55.06\\
Feature +Prototype & 55.04\\
 \rowcolor{mygray}
Combined +Prototype  &\textbf{54.97}\\
Combined+ Whole  &54.98\\
Combined +Random sample  &55.03\\ \hline
 Combined +Prototype+scratch &55.07\\\hline
\end{tabular}
}
\end{minipage}
\hspace{0.01\linewidth}
\begin{minipage}{0.29\linewidth}
\caption{\small{Test top-1 errors(\%) of ResNet-152 on Places-LT.
}}\label{placeslt}
\centering
\resizebox{1\textwidth}{!}
{
\begin{tabular}{c|c}
\toprule[1pt]
\textbf{Method} & \textbf{Places-LT}\\ \hline
CE& 69.3\\
Focal loss \cite{lin2017focal} (from \cite{liu2019large})& 65.4 \\
$\text{OLTR}$ \cite{liu2019large}&64.1\\
$\text{cRT}$ \cite{DBLP:conf/iclr/KangXRYGFK20} & 63.3\\
$\text{LWS}$ \cite{DBLP:conf/iclr/KangXRYGFK20} & 62.4\\
$\text{Mets-class-weight, CE}$\cite{jamal2020rethinking}&62.9\\
{BALMS \cite{ren2020balanced}}&{61.3}\\
{DisAlign\cite{zhang2021distribution}}&{60.7}\\
{$p_\text{meta}(y)/p_\text{train}(y)$}&{66.09}\\
{L2RW + RANDOM\cite{ren2018learning}}&{67.77}\\
 \rowcolor{mygray}
\textbf{Our method}&{\textbf{60.32$\pm$0.02}}\\
\bottomrule
\end{tabular}
}
\end{minipage}
\hspace{0.01\linewidth}
\begin{minipage}{0.24\linewidth}
\caption{\small{Test top-1 errors(\%) of ResNet-50 on ImageNet-LT. $*$ indicates results from  \cite{li2021metasaug}.}}\label{imagenet}
\centering
\resizebox{1.0\textwidth}{!}
{
\begin{tabular}{c|cc}
\toprule[1pt]
\textbf{Method} & \textbf{ImageNet-LT}\\ \hline
{CE}& 61.12\\
$\text{CB, CE}^*$\cite{cui2019class}&59.15 \\
$\text{OLTR}^*$\cite{liu2019large} &59.64\\
{$\text{LDAM}^*$ \cite{cao2019learning}}&58.14\\
{$\text{LDAM-DRW}^*$ \cite{cao2019learning}} & 54.26 \\
{$\text{Mets-class-weight, CE}^*$\cite{jamal2020rethinking}}&55.08\\
MetaSAug, CE  \cite{li2021metasaug}&52.61\\
\textbf{Our method+Reduced Prototype }&{52.41}\\
 \rowcolor{mygray}
\textbf{Our method}&{\textbf{52.36$\pm$0.01}}\\
\bottomrule
\end{tabular}
}
\end{minipage}
\vspace{-1mm}
\end{table}
To verify whether our method ameliorates the performance on minority classes, we plot the confusion matrices of CE, MetaSAug, and ours  on CIFAR-LT-10 with $\text{IF}\!=\! 200$ in Fig. \ref{fig:Confusion}. As expected, although CE training can almost perfectly classify the samples in majority classes, it suffers severe performance degeneration in the minority classes. MetaSAug improves the accuracies of the minority classes, where is still a big gap between the performance on the minority classes and the majority classes. In contrast, ours does not show a very clear preference for a certain class and outperforms the strong baseline on the overall performance, which is the goal of  on an imbalanced classification task.

\textbf{Experimental details and results on Places-LT and ImageNet-LT}
Following \citep{DBLP:conf/iclr/KangXRYGFK20}, we employ ResNet-152 pre-trained on the full ImageNet as the backbone on Places-LT. For stage 1, we set the initial learning rate as 0.01, which is decayed by $1e^{-1}$ every 10 epochs. In the stage 2 of our method, we only fine-tune the last fully-connected layer for training efficiency and set $\alpha$ as $1e^{-4}$ and $\beta$ as $1e^{-3}$ within 50 epochs. The mini-batch size is $32$ and the optimizer is SGD with momentum $0.9$ and weight decay $5e^{-3}$. As shown in Table \ref{placeslt}, our method outperforms all baselines. It further suggests that our method has excellent performance in the extreme imbalance setting with $\text{IF}\!=\!4980/5$.
For a fair comparison, we implement our method on ImageNet-LT with the same experimental conditions of \cite{li2021metasaug}, from which we have taken the results of other comparison methods. We consider ResNet-50 \cite{he2016deep} as the backbone on ImageNet-LT. In stage 1, we run 200 epochs and decay the learning rate by 0.1 at the 60th and 80th epochs. In stage 2, we implement our method for 50 epochs, set learning rate $\alpha$ as $2e^{-5}$ and $\beta$ as $1e^{-2}$, and only fine-tune the last fully-connected layer  for training efficiency. We use the SGD optimizer with momentum 0.9, weight decay $5e^{-4}$ and set the batch size as $128$. The results on ImageNet-LT of different models reported in Table \ref{imagenet} indicate the effectiveness of our proposed method on ImageNet-LT when comparing with strong baseline MetaSAug. Besides, we further consider randomly sampling a mini-batch of size 100 from all prototypes at each iteration to build $Q$, whose performance is comparable to the $Q$ from all prototypes. Thus, with a stochastic setting for $Q$, our proposed method can be used to the imbalanced training set with a large number of classes. {We defer the time computational complexity, additional quantitative results and qualitative results on different image datasets to Appendix \ref{sec:details_image}}.

\subsection{Experiments on  Imbalanced Text Classification}
\label{Experiments on  Imbalanced Text Classification}

\textbf{Datasets and settings}
Following \citep{hu2019learning, liu2021improving}, we adopt the popular SST-2 for 2-class and SST-5 for 5-class sentence sentiment \cite{socher2013recursive}. For a fair comparison, we use the same imbalanced datasets and settings with \citep{ liu2021improving}.
Specifically, we set class 1 as the minority class and the rest as the majority classes, where the number of examples in the majority class is fixed as 1000 (SST-2) and 500 (SST-5) and we achieve different imbalance settings by varying the number of examples in the minority class. Besides, the number of samples in the meta set is 10 for each class. We use the BERT (base, uncased) model \cite{devlin2018bert} as feature extractor and a simple 3-layer fully-connected network (FCN) with the structure in Appendix \ref{sec:details_text} as classifier. To make subsequent experiments on strong models, following \citep{liu2021improving}, we use an additional balanced training set (500 samples in each class) to fine-tune the BERT model, which is randomly selected from the remaining examples in each dataset except the imbalanced training set, meta set and to-be evaluated test set. Based on the fine-tuned BERT, we adopt the two-stage manner for the imbalanced text datasets, where we train the BERT + FCN in the first stage with CE loss and train the FCN with our proposed method by freezing the BERT in the second stage. The settings of the training process are deferred to Appendix \ref{sec:details_text}.


\textbf{Baselines} We consider the following methods: (1) \textbf{vanilla BERT}, the vanilla pretrained language model. (2) \textbf{Fine-tuned BERT }, where the pretrained BERT is fine-tuned on an additional balanced training set. (3) \textbf{Fine-tuned BERT + CE}, the fine-tuned BERT model followed by the FCN which is further trained by the CE loss on the imbalanced training set following \citep{hu2019learning, liu2021improving}. (4) \textbf{Automatic re-weighting methods}, including the method of \citet{hu2019learning} and constraint-based re-weighting \cite{liu2021improving}. Since few works consider imbalanced text classification, we further consider (5) \textbf{Empirical re-weighting methods}, including re-weighting with inverse class frequency ($i.e.$, Proportion) \cite{2016Learning,wang2017learning}) and LDAM-DRW \cite{cao2019learning} and  (6) \textbf{Logit adjustment} \cite{menon2020long} using their official codes and settings\footnote{\url{https://github.com/kaidic/LDAM-DRW}} \footnote{\url{https://github.com/google-research/google-research/tree/master/logit_adjustment}}. We repeat all experiments 10 times and report the mean and standard deviation.


\textbf{Experimental details and results on SST-2 and SST-5}
We report the text classification results of compared methods under different imbalance factors in Table \ref{sst2}. We find that our proposed method outperforms all competing methods in all imbalance factor settings, which demonstrates the effectiveness of our proposed method. Although all methods could achieve acceptable performance in a slight imbalance, the performance of three baselines (Vanilla BERT, Fine-Tuned BERT and Fine-Tuned BERT+CE) drop dramatically, indicating the importance of proposing specialized methods for handling imbalanced training datasets. Logit adjustment (post-hoc correction), is very competitive to ours on SST-2, which, however, only produces similar results to the three above-mentioned baselines on SST-5. In contrast, ours is robust to not only the imbalance factors but also the number of classes, where the results are consistent with the image case. We provide more results in Appendix \ref{text_ablation}. In addition to 1D text and 2D image, we further investigate the robustness of our method on 3D point cloud data, where we use the popular ModelNet10 \cite{wu20153d} and defer the experiments to Appendix \ref{sec:details_point}.


\begin{table}[!ht]
\centering
\caption{\small{Comparison of different models on SST-2 and SST-5. $\dag$ indicates results reported in \cite{liu2021improving}}.}
\resizebox{1\textwidth}{!}{
\begin{tabular}{c|cccc|ccc}
\toprule[1pt]
\multicolumn{1}{c|}{\textbf{Method}}&\multicolumn{4}{c|}{\textbf{SST-2}}&\multicolumn{3}{c}{\textbf{SST-5}}\\ \hline
\textbf{Imbalance Factor} & 1000 : 100 & 1000 : 50 &  1000 : 20 & 1000 : 10 & 500 : 75 & 500 : 60 & 500 : 50\\ \hline
Vanilla BERT&  74.91$\pm$4.62&53.26$\pm$5.70&50.54$\pm$1.40& 49.84$\pm$0.02&36.99$\pm$0.46&36.75$\pm$0.43&36.46$\pm$0.46\\
Fine-Tuned BERT &  81.64$\pm$3.79& 75.53$\pm$1.90&65.23$\pm$3.91&60.61$\pm$5.00& 43.76$\pm$0.77&43.25$\pm$0.73&42.70$\pm$0.52 \\
Fine-Tuned BERT+CE&  78.25$\pm$2.24& 57.18$\pm$1.88&55.00$\pm$1.23&50.17$\pm$1.34& 43.71$\pm$0.98&44.06$\pm$1.11&36.46$\pm$0.50\\
\hline
Proportion (reported by us) & 79.15$\pm$1.34&76.84$\pm$11.3&73.61$\pm$1.17 &69.52$\pm$14.4& 42.78$\pm$0.82 & 42.36$\pm$1.06 & 41.60$\pm$1.21\\
LDAM-DRW \cite{cao2019learning}(reported by us)&71.41$\pm$1.25&64.65$\pm$3.82&56.41$\pm$3.55&53.73$\pm$3.01&43.20$\pm$0.64&40.90$\pm$1.66&40.81$\pm$1.46\\
\hline
Hu et al.'s $^{\dag}$ \cite{hu2019learning} &81.57$\pm$0.74 &79.35$\pm$2.59 &73.61$\pm$11.9 &	55.84$\pm$11.8 &-  & - & 39.82$\pm$1.07\\
Hu et al.'s+Regularization $^{\dag}$ \cite{hu2019learning} &82.25$\pm$1.16 &-&79.53$\pm$1.64 &	66.68$\pm$14.0&-  &-  & 40.14$\pm$0.39 \\
Constraint-based re-weighting $^{\dag}$ \cite{liu2021improving}& 82.58$\pm$0.98 &-& 81.14$\pm$1.25& 80.62$\pm$0.93& -  & -  & 44.62$\pm$1.08\\
\hline
Logit Adjustment  \cite{menon2020long} (reported by us)& 86.37$\pm$0.30&86.61$\pm$0.31&86.51$\pm$0.33&86.50$\pm$0.38&43.52$\pm$2.63&39.52$\pm$2.03&36.55$\pm$2.77\\\hline
 \rowcolor{mygray}
\textbf{Our method (Weight Net)} & \textbf{87.08$\pm$0.09} & \textbf{87.13$\pm$0.04} & \textbf{87.14$\pm$0.08} & \textbf{87.10$\pm$0.05}&\textbf{44.95$\pm$0.56}&\textbf{44.79$\pm$0.82}&\textbf{44.68$\pm$0.98}\\
\bottomrule
\end{tabular}}\label{sst2}
\end{table}

\section{Conclusion}
This paper introduces a novel automatic re-weighting method for imbalance classification based on optimal transport (OT). This method presents the imbalanced training set as a to-be-learned distribution over its training examples, each of which is associated with a probability weight. Similarly, our method views another balanced meta set as a balanced distribution over the examples. By minimizing the OT distance between the two distributions in terms of the defined cost function, the learning of weight vector is formulated as a distribution approximation problem. Our proposed re-weighting method bypasses the commonly-used classification loss on the meta set and uses OT to learn the weights, disengaging the dependence of the weight learning on the concerned classifier at each iteration. This is an approach different from most of the existing re-weighting methods and may provide new thoughts for future work. Experimental results on a variety of imbalanced datasets of both images and texts validate the effectiveness and flexibility of our proposed method.


\clearpage
\bibliography{example_paper}
\bibliographystyle{unsrtnat}

\clearpage
\appendix

\section{Amortizing the learning of weight vector by introducing a weight net}

To integrate our proposed method with deep learning frameworks, we adopt a stochastic setting, $i.e.$, a mini-batch setting at each iteration. We adopt two-stage learning, where stage 1 trains the model $f(\thetav)$ by the standard cross-entropy loss on the imbalanced training set and stage 2 aims to learn the weight vector $\wv$ and meanwhile continue to update the model $f(\thetav)$.
Generally, at stage 2, calculating the optimal $\thetav$ and $\wv$ requires two nested loops of optimization, which is cost-expensive. We optimize $\thetav$ and $\wv$ alternatively, corresponding to \eqref{optimize_theta} and \eqref{OT_loss_final} respectively, where $\wv$ is maintained and updated throughout the training, so that  re-estimation from scratch can be avoided in each iteration. We summarize the amortized learning of $\wv$ in Algorithm \ref{Algorithm2}, where
the key steps are highlighted in Step (a), (b), and (c).
Specifically,
at each training iteration $t$, in Step (a), we have  $\hat{\thetav}^{(t+1)}(\Omegamat^{t})=\{\hat{\thetav}_1^{(t+1)}(\Omegamat^{t}),\hat{\thetav}_2^{(t+1)}(\Omegamat^{t})\}$ and $\alpha$ is the step size for $\thetav$; in Step (b), as the cost function based on features is related with $\hat{\thetav}_1^{(t+1)}(\Omegamat^{t})$, the OT loss relies on $\hat{\thetav}_1^{(t+1)}(\Omegamat^{t})$, and $\beta$ is the step size for $\Omegamat$; in Step (c), we ameliorate model parameters $\thetav^{(t+1)}$.

\begin{algorithm}[]
\footnotesize
\caption{\small{{Workflow about our re-weighting method for optimizing $\thetav$ and weight net.}}}
\begin{algorithmic}
 \STATE \textbf{Require:} Datasets $\mathcal{D}_{\text {train }}$, $\mathcal{D}_{\text {meta }}$, initial model parameter $\thetav$ and weight net $\Omegamat$, hyper-parameters $\{\alpha,\beta,\lambda\}$
\FOR{$t=1,2,...,t_1$}
  \STATE Sample a mini-batch $B$ from the training set $\mathcal{D}_{\text {train }}$;
  \STATE Update $\thetav^{(t+1)} \leftarrow \thetav^{(t)}-\alpha \nabla_{\thetav} \mathcal{L}_{B}$ where $\mathcal{L}_{B}=\frac{1}{|B|} \sum_{i \in B} \ell\left(y_{i}, f\left(x_{i} ; \thetav^{(t)} \right)\right) $;
\ENDFOR
\FOR{$t=t_1+1,...,t_1+t_2$}
  \STATE Sample a mini-batch $B$ from the training set $\mathcal{D}_{\text {train }}$;
    \STATE Compute weight vector $\wv^{t}$ by weight net $\Omegamat^{t}$;
  \STATE \textbf{Step (a):} Update $\hat{\thetav}^{(t+1)}(\wv^{(t)}) \gets \thetav^{(t)}-\alpha \nabla_{\thetav} \mathcal{L}_{B}$ where $\mathcal{L}_{B}=\frac{1}{|B|} \sum_{i \in B} w_i^{(t)} \ell\left(y_{i}, f\left(x_{i} ; \thetav^{(t)} \right)\right)$
      \STATE Use $\mathcal{D}_{\text {meta }}$ to build $Q$ in \eqref{new_Q_distribution} and $B$ with $\wv^{t}$ to build  $P(\wv^{t})$ \eqref{P_distribution};
  \STATE \textbf{Step (b):} Compute $L_{\text{OT}} \left(\hat{\thetav}_1^{(t+1)}(\Omegamat^{t}),\Omegamat^{t}\right)$ with cost \eqref{OT_sink_combine}; Optimize $\Omegamat^{t+1} \gets \Omegamat^{t}-\beta \nabla_{\Omegamat} L_{\text{OT}} \left(\hat{\thetav}_1^{(t+1)}(\Omegamat^{t}),\Omegamat^{t}\right)$
  \STATE \textbf{Step (c):} Update ${\thetav}^{(t+1)}\gets \thetav^{(t)}-\alpha \nabla_{\thetav} \mathcal{L}_{B}$ where  $\mathcal{L}_{B}=\frac{1}{|B|} \sum_{i \in B} w_i^{(t+1)} \ell\left(y_{i}, f\left(x_{i} ; \thetav^{(t)} \right)\right) $ and we compute $w_i^{(t+1)}$ with updated weight net $\Omegamat^{t+1}$
  \ENDFOR
\end{algorithmic}\label{Algorithm2}
\end{algorithm}

\section{More details and results about the imbalanced image classification}\label{sec:details_image}

\subsection{Details about imbalanced training datasets}
\textbf{\textit{CIFAR-LT}} is the long-tailed version of CIFAR dataset, where the original CIFAR-10 (CIFAR-100) dataset \cite{krizhevsky2009learning} has 5000 (500) images per class and falls into 10 (100) classes. Following \cite{li2021metasaug}, we create long-tailed training sets from CIFAR-10 and CIFAR-100 by discarding some training samples, where we vary the imbalance factor $\textrm{IF}\in\{200, 100, 50, 20\}$. We do not change the balanced test sets and we randomly select ten training images per class as our meta set.

We build \textbf{\textit{ImageNet-LT}} based on the classic ImageNet \cite{deng2009imagenet} following \cite{liu2019large}. After discarding some training examples, ImageNet-LT has 115.8K training examples in 1,000 classes, with an imbalance factor $\textrm{IF}=1280/5$. The authors also provided a small balanced validation set with 20 images per class, from which we sample 10 images to construct our meta set as \cite{li2021metasaug}. Besides, the original balanced ImageNet validation set is adopted as the test set (50 images per class).

\textbf{\textit{Places-LT}} is created from Places-2 \cite{zhou2017places} by \cite{liu2019large} with the same strategy as above. The Places-LT contains 62.5K images from 365 categories with $\textrm{IF}=4980/5$. This factor means that Places-LT is more changeling than ImageNet-LT dataset. Similarly, there is an official balanced validation dataset with 20 images per class, where we random select 10 images to build our meta set following \cite{jamal2020rethinking}.

\textbf{\textit{iNaturalist 2018}} \footnote{\url{https://github.com/visipedia/inat_comp/tree/master/2018}} is collected from real world which contains 435,713 training samples in 8142 categories with the imbalance factor $\textrm{IF}$ of $1000/2$. Following \citep{li2021metasaug, jamal2020rethinking}, we use the original validation dataset to evaluate our method and random select 2 samples from every category in training dataset to construct our meta set.

\subsection{Experimental details and results on iNaturalist 2018}
Following \citep{jamal2020rethinking}, we use ResNet-50 pre-trained on ImageNet plus iNaturalist 2017 as the backbone network on iNaturalist 2018. For stage 1, we set the initial learning rate as 0.01, which is decay by $1e^{-1}$ every 20 epochs. In the stage 2 of our method, we only fine-tune the last fully connected layer and set $\alpha$ as $1e^{-4}$ and $\beta$ as $1e^{-3}$ within 200 epochs. The mini-batch size is 64, the optimizer is SGD with momentum $0.9$ and weight decay is $5e^{-3}$.

Considering the label vector from the iNaturalist 2018 dataset is high-dimensional and sparse, we only use the features of samples to define the cost function, $i.e.$, \textit{Feature-aware cost}. For training efficiency, we use \textit{Prototype} to construct our meta set. As summarized in Table \ref{ina2018}, our proposed method outperforms all the baselines. This observation demonstrates that our proposed re-weighting method can effectively deal with the extremely imbalanced large-scale training set with a large number of classes.


\subsection{Learned weights for different classes with varying imbalanced settings}
\begin{figure}[h]
 \setlength{\abovecaptionskip}{-0.1cm}
\setlength{\belowcaptionskip}{-0.1cm}
 \centering
\includegraphics[height=6cm]{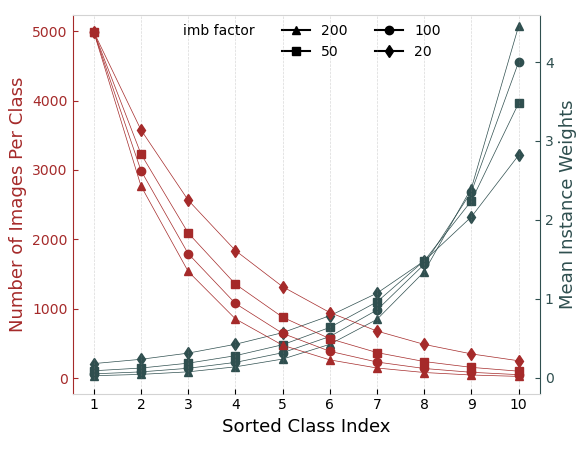}
 \caption{\small{The number of samples and learned mean weights within each class with varying $\text{IF}$ on CIFAR-LT-10.}}\label{fig:class_weight}
\end{figure}

To explore the learned example weights, we average them within each class and visualize the averaged weight for each class with varying imbalanced factors in Fig. \ref{fig:class_weight}. We see that the learned weights of the class 10 (tail class) are more prominent than those of the majority classes. This phenomenon becomes increasingly evident as the training set becomes more imbalanced.


\subsection{Convergence and time complexity}
In this section, we investigate the convergence and time complexity. {The optimization of the OT distance between discrete distributions of size $n$, is a linear program and can be solved with combinatorial algorithms such as the simplex methods, which dampens the utility of the method when handling large datasets. Recently, the regularization scheme proposed by Cuturi in \cite{cuturi2013sinkhorn} allows a very fast computation of a transportation plan, denoted as Sinkhorn algorithm. The Sinkhorn algorithm requires the computational cost of $n^{2} \log (n)/ \varepsilon^{2}$ to reach $\varepsilon$-accuracy, which is adopted by us to solve the OT problem.}

\begin{figure}
 \setlength{\abovecaptionskip}{-0.1cm}
\setlength{\belowcaptionskip}{-0.1cm}
 \centering
\includegraphics[height=4.7cm]{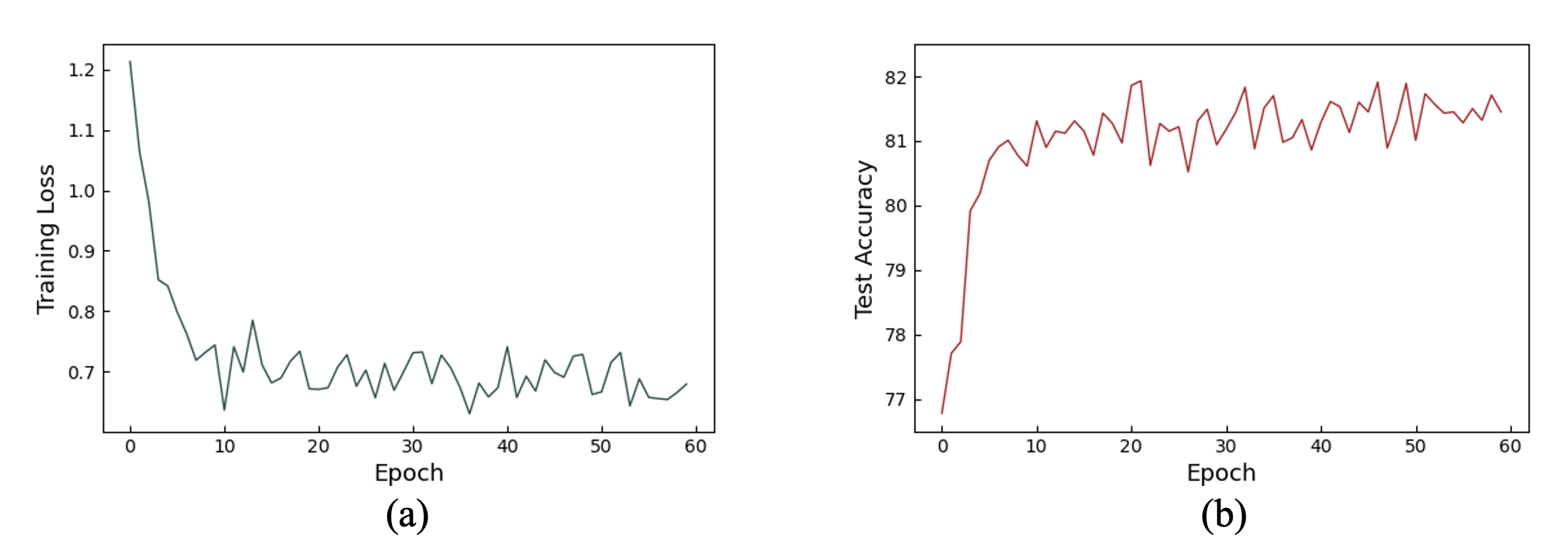}
 \caption{\small{Imbalanced training loss (a) and balanced test accuracy (b) on CIFAR-10 with imbalanced factor 100.}}\label{fig:convergence}
\end{figure}

As shown in Fig. \ref{fig:convergence}, we plot the variation of training loss and test classification performance as the increase of training time at the second stage. The training loss of our proposed method can converges at around 10th epoch, where 20 epochs are usually enough for our method to yield satisfactory performance. This means our method can reach fast convergence.

To investigate the time complexity, we conduct experiments on CIFAR 10, CIFAR100, ImageNet-LT and Places-LT with different imbalanced factor for ten epoches and average over the time. Models on CIFAR are evaluated with batch size of 16 on a single RTX 3080 GPU. Models on ImageNet-LT are evaluated with batch size of 128 on 4 RTX 2080Ti GPU. Models on Places-LT and iNaturalist 2018 are evaluated with batch size of 32 and 64 on 4 Tesla V100 GPU, respectively. The results are summarized in Table \ref{time}. We observe that our proposed method has an acceptable time complexity compared with the cross entropy. It means that our proposed method achieves a better performance with an acceptable cost by introducing the OT loss for re-weighting.



\begin{table}\small
\centering
\vspace{-3mm}
\caption{\small{Test top-1 errors(\%) of ResNet-50 on iNaturalist 2018. $*$ indicates results from \cite{li2021metasaug}.}}
\resizebox{0.4\textwidth}{!}{
\begin{tabular}{c|c}
\toprule[1pt]

\textbf{Method} & \textbf{iNaturalist 2018}\\ \hline
{CE}& 34.24\\
$\text{CB, CE}^*$\cite{cui2019class}&33.57 \\
{$\text{LDAM}^*$ \cite{cao2019learning}}&34.13\\
{$\text{LDAM-DRW}^*$ \cite{cao2019learning}} & 32.12 \\
$\text{cRT}^*$ \cite{DBLP:conf/iclr/KangXRYGFK20} & 32.40\\
{$\text{Mets-class-weight, CE}^*$\cite{jamal2020rethinking}}&32.45\\
{$\text{MetaSAug, CE}^*$  \cite{li2021metasaug}}&31.25\\
 \rowcolor{mygray}
\textbf{Our method}&\textbf{30.59}\\
\bottomrule
\end{tabular}}\label{ina2018}
\end{table}

\begin{table}
\centering
\vspace{-3mm}
\caption{\small{Running time(s) of our method and the baseline method on CIFAR-LT-10, CIFAR-LT-100, ImageNet-LT, Places-LT and iNaturalist (iNat) 2018 under different settings.}}
\resizebox{1.0\textwidth}{!}{
\begin{tabular}{c|cccc|cccc|c|c|c}
\toprule[1pt]
\multicolumn{1}{c|}{\textbf{Datasets}}&\multicolumn{4}{c|}{\textbf{CIFAR-LT-10}} &\multicolumn{4}{c|}{\textbf{CIFAR-LT-100}}&\multicolumn{1}{c|}{\textbf{ImageNet-LT}}&\multicolumn{1}{c|}{\textbf{Places-LT}}&\multicolumn{1}{c}{\textbf{iNat 2018}} \\\hline
\textbf{Imbalance Factor} & 200 & 100 & 50 & 20&200&100&50&20&1280/5&4980/5&1000/2 \\ \hline
{Cross-entropy (CE)}&  11.68 & 13.06 & 14.91 & 18.22 & 9.70& 11.53 &13.03 &16.50&241.46&375.46&3813.22\\ \hline
{Our method (Weight Vector)}&{13.70} & {15.26} &{ 17.36}& 21.18&{11.56}&{13.43}&{15.44}&19.95&252.16&381.09&4237.73\\
\bottomrule
\end{tabular}}\label{time}
\end{table}

\subsection{Learned Transport plan matrix and weight vector}
\begin{figure}
\begin{minipage}{.49\linewidth}
 \setlength{\abovecaptionskip}{-0.1cm}
\setlength{\belowcaptionskip}{-0.1cm}
 \centering
\includegraphics[height=6.9cm]{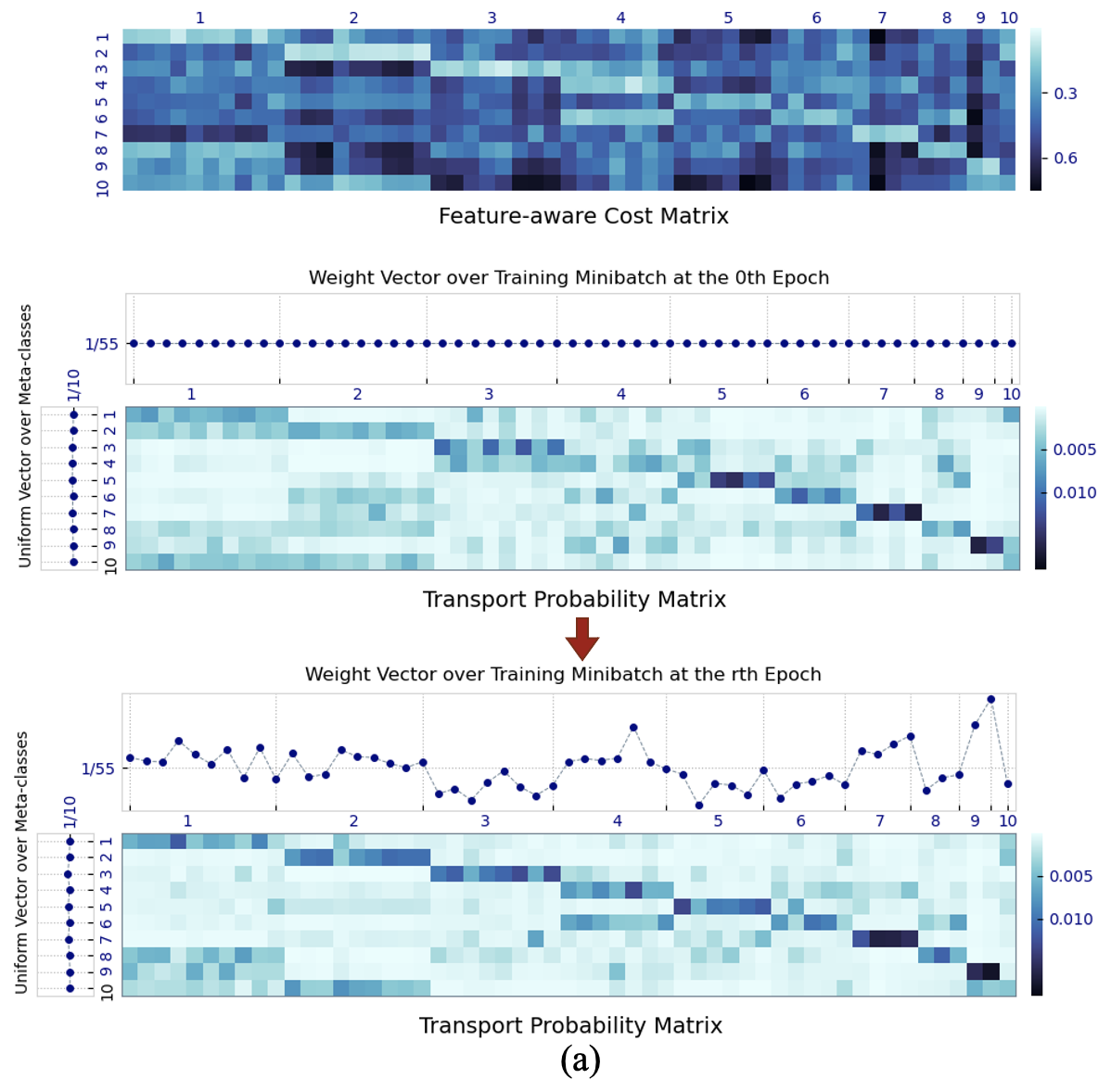}
 \end{minipage}
\hspace{0.01\linewidth}
\begin{minipage}{.49\linewidth}
 \setlength{\abovecaptionskip}{-0.1cm}
\setlength{\belowcaptionskip}{-0.1cm}
 \centering
\includegraphics[height=6.9cm]{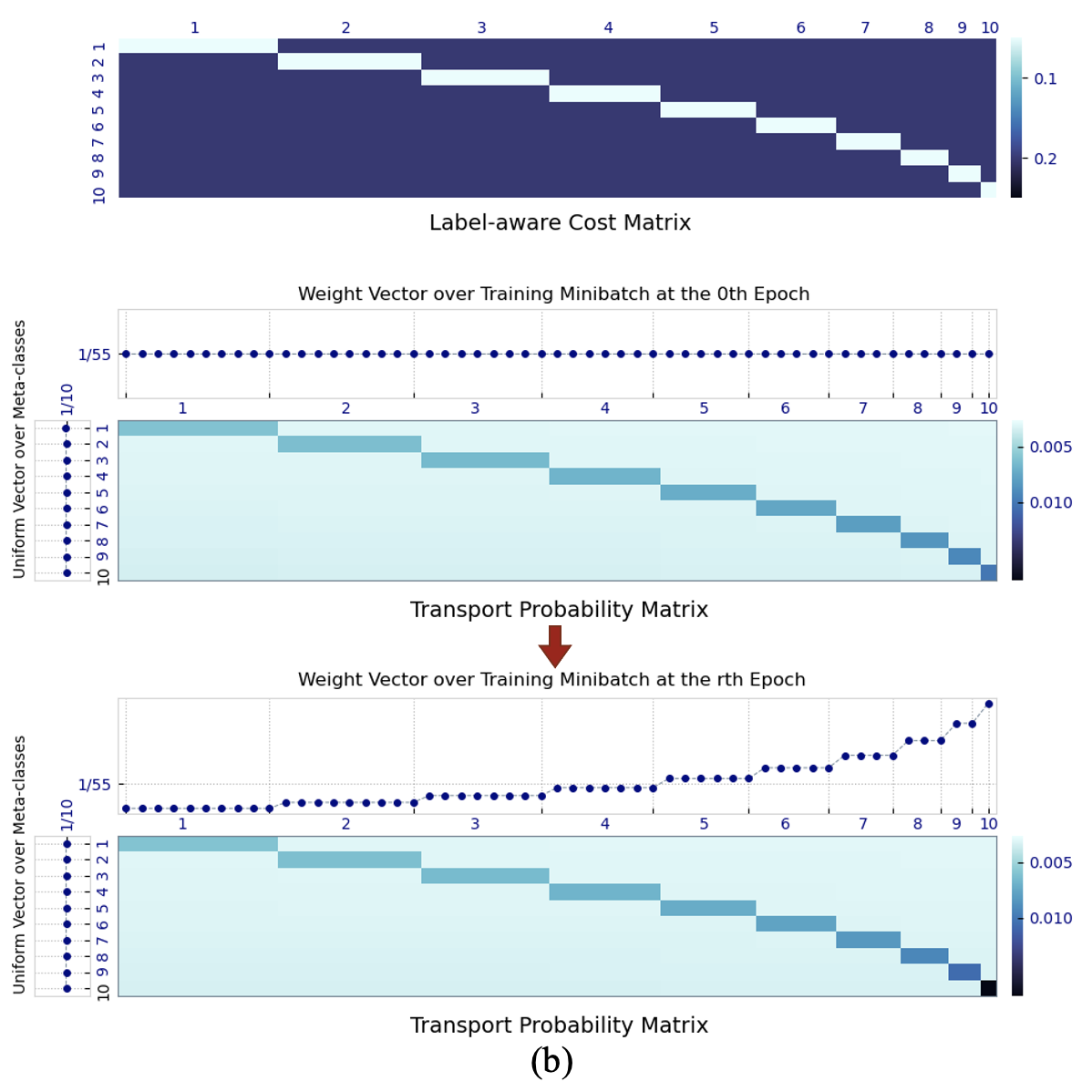}
 \end{minipage}
 \begin{minipage}{.49\linewidth}
 \setlength{\abovecaptionskip}{-0.1cm}
\setlength{\belowcaptionskip}{-0.1cm}
 \centering
\includegraphics[height=7.2cm]{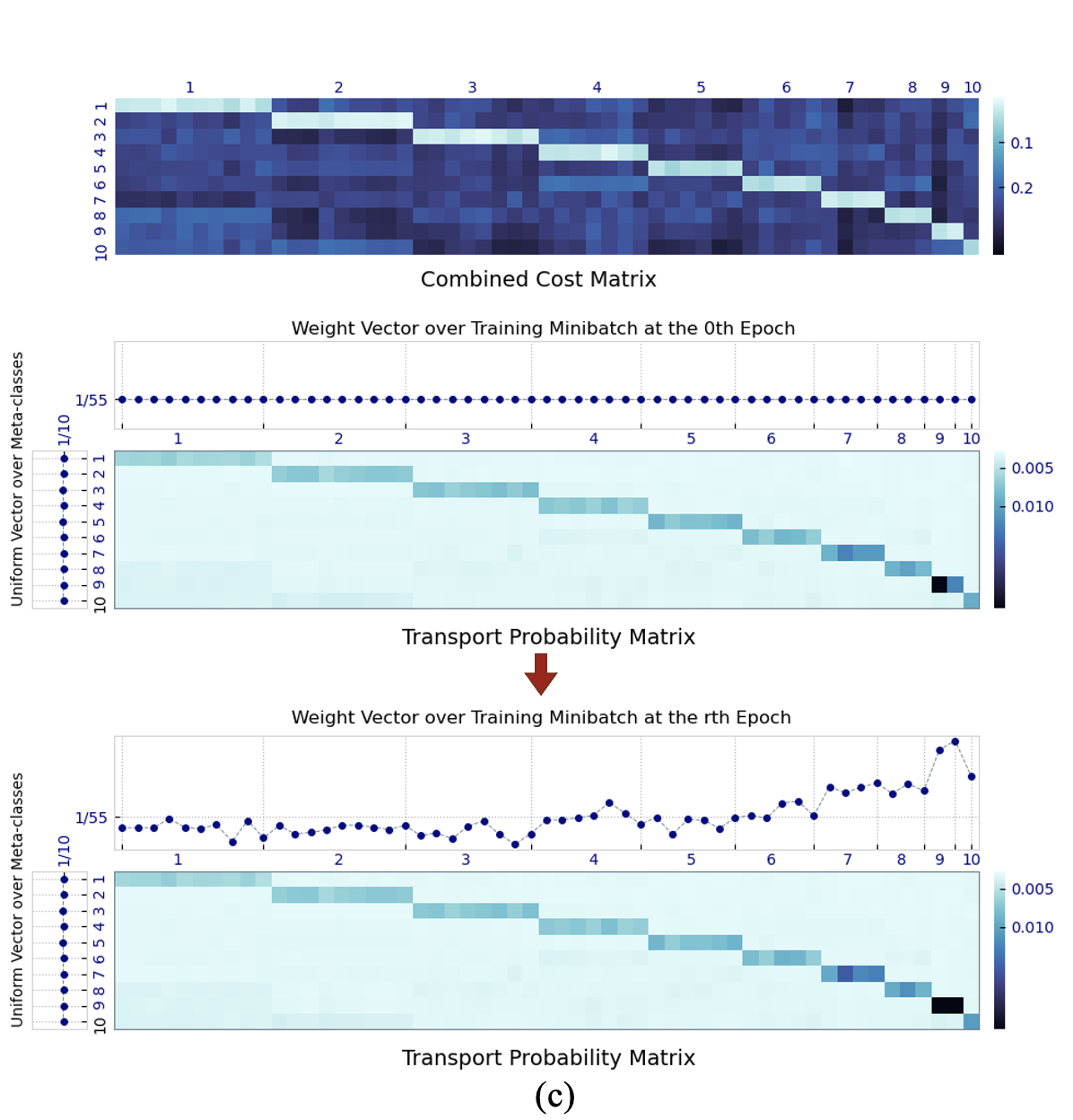}
 \end{minipage}
 \hspace{0.06\linewidth}
 \begin{minipage}{.4\linewidth}
 \setlength{\abovecaptionskip}{-0.1cm}
\setlength{\belowcaptionskip}{-0.1cm}
 \centering
\caption{\small{The process of learning the optimal transport given feature-aware cost matrix, label-aware cost matrix and combined cost matrix respectively. In each subfigure, the top row is the fixed cost matrix if the feature extractor is frozen. The middle row is the initialization of the training set(top) and the meta set(left) probability measure (both uniform), and the coupling transport probability between them. The bottom row is the learned training set measure (i.e. weight vector), meta set measure and according transport probability matrix after training for $r$ epochs.}}\label{fig:toy example}
\centering
\end{minipage}
\end{figure}

To explain the effectiveness of learning the weight vector with OT, we give a toy example. The imbalanced training mini-batch consists of $\{10,9,...,1\}$ training samples from class $\{1,2,...,10\}$ and the balanced meta set consists of $\{10,10,...,10\}$ validation samples from class $\{1,2,...,10\}$, which are used to compute the prototype for each class. The process of learning the weight vector is shown in Fig. \ref{fig:toy example}, where we use feature-aware cost, label-aware cost and combined cost in (a)-(c), respectively. For all experiments, we fix the probability measure of meta set as uniform distribution, initialize the to-be-learned probability measure ($i.e.$, $\wv$) of training mini-batch with a uniform measure, and we then optimize $\wv$ by minimizing the OT loss with given cost function. Given different cost functions, we find that the learned weight vectors and transport probability matrices show different characteristics. Given the feature-aware cost function, we can assign weight to examples instance-wisely. However, the inaccurate discrimination of features may mislead the learning of weight. Under the label-aware cost function, the learned weights are class-level and inversely related to the class frequency. Given the combined cost function, the results combine the benefits of label-aware and feature-aware cost functions. Interestingly, the weights assigned to examples of tail classes are more prominent than those of the head classes, where the weights of head classes are below the initialization $1/55$ and those of tail classes are above $1/55$. Such results demonstrate our re-weighting method can pay more attention to the tail classes.

\begin{figure}
 \centering
\includegraphics[height=4.1cm]{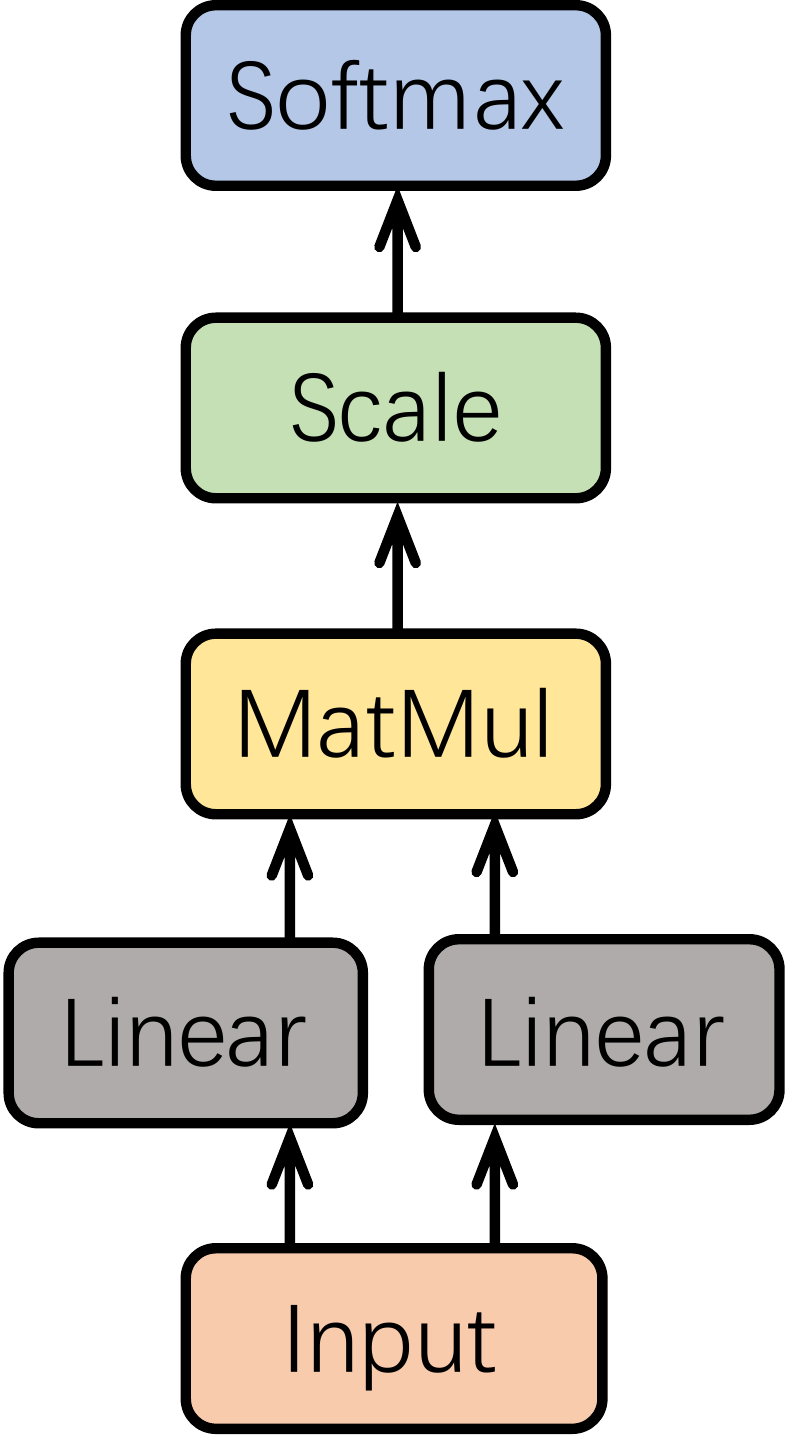}
 \caption{The weight net structure used in text classification experiment.}\label{fig:weightnet}
\end{figure}

\begin{table}
\caption{Hidden shape in weight net. What we feed into the net is the output vector from feature extractor in backbone network. We give a detailed shape information in weight net when assuming $d_l$ is 128 and $d_f$ is $[32, 768]$, where 32 represents batch size.}\label{tab:weightnet}
\centering
\resizebox{0.2\textwidth}{!}
{
\begin{tabular}{c|c}
\toprule[1pt]
Layer &  Shape\\ \hline
Input& [32, 768] \\
Linear& [32, 128]\\
MatMul& [32, 32]\\
Scale& [32, 32]\\
Softmax& [32, 32]\\ \hline
\end{tabular}
}
\end{table}

\section{More details about the text classification}\label{sec:details_text}

\subsection{FCN structure}
We give the FCN structure in Table \ref{Tab:structure}.

\subsection{Weight net structure on text classification task}\label{sec:weightnet_structure}
We give the weight net structure used in text classification task in Fig. \ref{fig:weightnet}. Different that used in image and point cloud classification tasks, we build a self attention-similar weight net and take the sample features $z_i^{\text{train}}$ as input:
\begin{align}\label{selfatten}
\wv =\operatorname{softmax}\left(\sv/\sqrt{d_l}\right), s^{i} \!=\! \left(\Wmat_{1}\zv_{i}^{\text{train}}\right)\left(\Wmat_{2} \zv_{i}^{\text{train}}\! \right)^T ,
\end{align}
Specifically, the output of dimension $d_f$ from feature extractor will be encoded to the same dimension $d_l$ by two separate linear layers $\Wmat_{1}$ and $\Wmat_{2}$. We compute their dot products, divide by a scaled factor $\sqrt{d_l}$ and apply a softmax operator to obtain expected instance weights. The hidden shape in our experiments is shown in Table \ref{tab:weightnet}.

\begin{table}
\caption{\small{The network structure for 3-layer FCN on text classification experiments. \textbf{cls\_num} is 2 for SST-2 and 5 for SST-5.}}
\centering
\begin{tabular}{c|c}
\toprule[1pt]
&Input 768-dimensional text features from BERT \\ \hline
Feature extractor & Linear layer (768, 768), ReLU \\
Feature extractor & Linear layer (768, 768), ReLU \\
Classifier & Linear layer (768, cls\_num), Tanh \\
\bottomrule
\end{tabular}\label{Tab:structure}
\end{table}

\begin{table}
\caption{\small{Settings of the training on SST-2 and SST-5 dataset.}}
\centering
\begin{tabular}{c|ccc}
\toprule[1pt]
Dataset&Fine-tune / Pretrain & Stage 1 & Stage 2 \\ \hline
&Adam ($5e^{-6}$)&SGD ($1e^{-4}$) & SGD ($\alpha~1e^{-4}$) and SGD ($\beta~4e^{-2}$) \\
SST-2 or SST-5& epochs:5 & epochs:15 & epochs:10 \\
& batch size:8 & batch size:50 & batch size:50 \\
\bottomrule
\end{tabular}\label{Tab:setting_text}
\end{table}

\subsection{Experimental settings}
For fining tune BERT, we set the initial learning rate as $5e^{-6}$, where epochs is 5 and batch size is 8. For the first stage, we use SGD optimizer with learning rate $1e^{-4}$. For the second stage, we use SGD optimizer with $\alpha$ learning rate $1e^{-4}$ and $\beta$ learning rate $4e^{-2}$. For both two stages, our minibatch size is 50 and epochs are 15 and 10, respectively. The settings of the training process on the STT-2 and SST-5 are listed in Table \ref{Tab:setting_text}.

\subsection{Ablation Study}\label{text_ablation}
To prove the robustness of our method and its variants on text data, we execute an ablation study on SST-2 dataset, which is shown in Table \ref{Tab:sst2_ablation}. We can find that combined cost achieves best performance and using label or feature cost can still obtain an acceptable result, which indicates the OT cost is flexible and robustness. When comparing the influence of different ways in constructing meta set, we can see that both whole meta set and prototype-based meta set achieve a close performance, while random sample meta cause a noticeable performance drop.

\begin{table}[!ht]
\centering
\caption{Ablation study on SST-2 with learned weight net under different imbalance factors.}
\resizebox{0.8\textwidth}{!}{
\begin{tabular}{c|cccccc}
\toprule[1pt]
\textbf{Method} & 100 : 1000 & 50 : 1000 & 20 : 1000 & 10 : 1000 \\ \hline
\textbf{Label + Whole}& 86.48$\pm$0.29&86.30$\pm$0.37 & 86.19$\pm$0.54 &86.03$\pm$0.46 \\
\textbf{Feature + Whole}&86.08$\pm$0.38 &86.08$\pm$0.37 &85.98$\pm$0.29  &85.95$\pm$0.47 \\
\textbf{Combined + Whole}& 86.71$\pm$0.24 & 86.49$\pm$0.60 & 86.47$\pm$0.51 &86.24$\pm$0.60\\ \hline
\textbf{Combined + Whole}& 87.08$\pm$0.09 & 87.13$\pm$0.04 & 87.14$\pm$0.08 & 87.10$\pm$0.05\\
\textbf{Combined + Prototype}& 87.12$\pm$0.03 & 87.11$\pm$0.01 & 87.09$\pm$0.07 & 87.12$\pm$0.01\\
\textbf{Combined + Random sample}& 86.59$\pm$0.50 & 86.14$\pm$0.52 & 86.08$\pm$0.77 & 85.85$\pm$0.52\\
\bottomrule
\end{tabular}}\label{Tab:sst2_ablation}
\end{table}

\section{More details about the point cloud classification}\label{sec:details_point}

\subsection{Experiments on Imbalanced Point Cloud Classification}

\textbf{Dataset and Baselines} In addition to 1D text and 2D image, we further investigate the robustness of our method on 3D point cloud data, where we use the popular ModelNet10 \cite{wu20153d}.
We select PointNet++ \cite{qi2017pointnet++} as our backbone without other additional layers. We use 5  examples in each class for meta set. Specifically, we down-sample the original points to 512 followed by a random point dropout, scale and shift to process the input data, which is a common setting for point cloud classification. We conduct the same process on all the experiments to evaluate our method fairly. Since we are the first to conduct imbalance classification task on point cloud data, we consider following methods: (1) \textbf{Pointnet++ baseline\footnote{\url{https://github.com/yanx27/Pointnet_Pointnet2_pytorch}}}, a classic network used in point cloud classification \cite{qi2017pointnet++}. (2) \textbf{Focal loss} \cite{lin2017focal}. (3) \textbf{LDAM-DRW} \citep{cao2019learning}. (4) \textbf{Logit Adjustment} \cite{menon2020long}.

\textbf{Experimental details and results on ModelNet10} The classification results of different methods are summarized in Table \ref{Tab:modelnet}, where we evaluate our method from the aspects of instance accuracy (IA) and class accuracy (CA). We can see that ours outperforms all baselines on IA and CA. These experimental results further demonstrate the effectiveness of integrating OT loss into imbalanced point cloud classification, showing the generalization ability of our proposed method to data modalities.

\subsection{Experimental settings}
 For point cloud classification, we select \textit{bathtub} as minority class and the rest as majority, where the number of training sample is 100 in every class. We train the baseline model from scratch where we use Adam optimizer with learning rate $1e^{-4}$. Besides, our minibatch size is 128 and epochs are 40. For the first and second stage, we still use Adam optimizer, where learning rate is $1e^{-4}$ in the 1-st stage, $\alpha$ learning rate $1e^{-4}$ and $\beta$ learning rate $5e^{-3}$. For both two stages, we keep the same number of epochs and minibatch size as them in training the baseline. For training and meta dataset, we choice 100 samples from every majority class in original training set, where we then random choice 5 samples from the rest training set to build our meta set. For test dataset, we random sample 40 instances in every class from the original test dataset to construct our balanced test dataset.
 \begin{table}[h]
\centering
\caption{Comparison of the proposed method and {baselines} trained with different proportions on ModelNet10 for multi-class classification.}
\resizebox{0.95\textwidth}{!}{
\begin{tabular}{c|ccccccccc}
\toprule[1pt]
\multirow{2}{*}{\textbf{Method}} & \multicolumn{2}{c}{1:100} & \multicolumn{2}{c}{4:100} & \multicolumn{2}{c}{8:100} & \multicolumn{2}{c}{10:100} \\
                        & IA          & CA          & IA          & CA          & IA          & CA          & IA           & CA          \\ \hline
Pointnet++ Baseline\cite{qi2017pointnet++}                & 81.75$\pm$0.82           & 80.59$\pm$1.07           & 82.53$\pm$1.10           & 81.42$\pm$1.35           & 88.67$\pm$0.37           & 88.28$\pm$0.75           & 87.95$\pm$0.27            &87.58$\pm$0.42            \\\hline
Focal Loss $\alpha=0.25$, $\gamma=2$ \cite{lin2017focal}                & 82.09$\pm$0.68           & 80.74$\pm$0.68           & 87.17$\pm$0.71           &86.39$\pm$1.01           & 87.44$\pm$1.43            &87.49$\pm$0.43  & 87.67$\pm$0.78           & 87.11$\pm$0.21                       \\
LDAM-DRW \cite{cao2019learning}              & 79.91$\pm$0.76           & 77.56$\pm$1.13           & 80.07$\pm$1.15           &78.19$\pm$1.29           & 80.35$\pm$0.93            &77.65$\pm$0.93  & 80.13$\pm$0.31           & 77.61$\pm$2.76                       \\\hline
Logit Adjustment $\tau=1.0$ \cite{menon2020long}                & 82.81$\pm$0.57           & 81.86$\pm$0.41           & 84.88$\pm$0.93           &\textbf{84.39$\pm$1.28}           & 86.77$\pm$1.07            &86.28$\pm$0.88  & 87.56$\pm$1.13           & 87.13$\pm$1.20                       \\
 \rowcolor{mygray}
\textbf{Our method (Weight vector)}                     & \textbf{83.71$\pm$0.11}           & \textbf{81.97$\pm$1.59}           & \textbf{85.15$\pm$1.01}           & 83.80$\pm$0.54           & \textbf{89.73$\pm$0.32}           & \textbf{89.09$\pm$0.38}           & \textbf{90.40$\pm$0.17}           & \textbf{90.16$\pm$0.56}           \\
\hline
\end{tabular}}\label{Tab:modelnet}
\end{table}

\begin{table}[h]
\caption{\small{Settings of the training on ModelNet10 dataset.}}
\centering
\begin{tabular}{c|cc}
\toprule[1pt]
Dataset & Stage 1 & Stage 2 \\ \hline
&Adam ($1e^{-4}$)& Adam ($\alpha~1e^{-4}$) and Adam ($\beta~5e^{-3}$) \\
ModelNet10 & epochs:10 & epochs:50 \\
 & batch size:64 & batch size:64 \\
\bottomrule
\end{tabular}\label{Tab:setting_point}
\end{table}

\begin{table}[htbp!]
\centering
\caption{{Ablation study on ModelNet 10 under different imbalance factors.}}
\resizebox{1.0\textwidth}{!}{
\begin{tabular}{c|ccccccccc}
\toprule[1pt]
\multirow{2}{*}{\textbf{Method}} & \multicolumn{2}{c}{1:100} & \multicolumn{2}{c}{4:100} & \multicolumn{2}{c}{8:100} & \multicolumn{2}{c}{10:100} \\
                        & IA          & CA          & IA          & CA          & IA          & CA          & IA           & CA          \\ \hline

\textbf{Weight Vector + Label + Whole}                     & 81.36$\pm$0.90          & 80.13$\pm$0.77           & 82.53$\pm$1.00           & 81.61$\pm$1.07           & 86.95$\pm$0.88           & 86.35$\pm$1.02           & 87.33$\pm$0.62          & 86.76$\pm$0.67           \\

\textbf{Weight Vector + Feature + Whole}                     & 81.81$\pm$0.86          & 80.59$\pm$0.87           & 83.20$\pm$0.40           & 82.16$\pm$0.34           & 87.00$\pm$0.83           & 86.35$\pm$0.64           & 88.06$\pm$0.40          & 87.43$\pm$0.14           \\
\textbf{Weight Vector + Combined + Prototype}                     & 83.04$\pm$0.35           & 81.97$\pm$0.32           & 85.16$\pm$1.28           & 84.11$\pm$1.10           & 89.73$\pm$0.69           & 89.46$\pm$0.64           & 89.84$\pm$0.19           & 89.50$\pm$0.41           \\
\textbf{Weight Vector + Combined + Whole}                     & 83.71$\pm$0.11          & 81.97$\pm$1.59           & 84.15$\pm$1.01           & 83.80$\pm$0.54           & 89.73$\pm$0.32           & 89.09$\pm$0.38           & 90.40$\pm$0.17          & 90.16$\pm$0.56           \\

\textbf{Weight Net + Combined + Whole}                & 82.81$\pm$0.63           & 81.59$\pm$1.07           & 83.37$\pm$0.33           & 81.89$\pm$0.41           & 89.56$\pm$0.48           & 88.81$\pm$0.86           & 89.84$\pm$0.73            &88.86$\pm$0.46            \\
\hline
\end{tabular}}\label{modelnet_ablation}
\end{table}
\subsection{Ablation study}
Similarly, we execute an ablation on point cloud to investigate the influence of OT loss, which is shown Table \ref{modelnet_ablation}. We can see that the best performance is obtained when we use the setting of weight vector, combined cost and whole meta set. Besides, performance are also acceptable when we use label cost and feature cost. The performance is also competitive when we use the setting of weight vector, combined cost and prototype-based meta set.

\subsection{Negative Societal Impacts}
This paper introduces a re-weighting method based on optimal transport to solve the imbalanced problem. The proposed re-weighting method is very different from existing re-weighting methods and may have the potential to provide new and better thoughts for imbalanced problems in the future. Notably, once a malicious imbalanced task is selected, our re-weighting method may produce a negative societal impact, similar to the progress in other machine learning methods.

{\section{Effect of Different Meta-sets }}\label{sec:dynamic_meta_exp}
{To explore whether our proposed method is effective even without an additional meta set from the validation set, we build a dynamic meta set from the given imbalanced training set. Notably, ``dynamic'' means that we randomly choose the samples from the training set to build a meta set at each training stage. Due to the limited rebuttal time, we report experiment results in SST-2 dataset in Table \ref{sst2_metaset}. The results show our method could still reach a comparable performance even with the meta set built from the imbalance training dataset dynamically.}

\begin{table}[!ht]
\centering
\caption{{Comparison of different meta set in SST-2 dataset, where additional meta set os built from the validation dataset and dynamic meta set is built from the imbalance training dataset directly.}}
\resizebox{1.0\textwidth}{!}{
\begin{tabular}{c|cccccc}
\toprule[1pt]
\textbf{Method} & 100 : 1000 & 50 : 1000 & 20 : 1000 & 10 : 1000 \\ \hline
\textbf{BERT Baseline (Without Fine Tuned)}&  74.91$\pm$4.62&53.26$\pm$5.70&50.54$\pm$1.40& 49.84$\pm$0.02\\
\textbf{BERT Baseline (Fine Tuned)}&  81.64$\pm$3.79& 75.53$\pm$1.90&65.23$\pm$3.91&60.61$\pm$5.00\\
\textbf{Baseline}&  78.25$\pm$2.24& 57.18$\pm$1.88&55.00$\pm$1.23&50.17$\pm$1.34\\
\textbf{Proportion  $\dag$} & 79.59$\pm$3.35&\XSolidBrush&78.76$\pm$2.40 &60.63$\pm$13.13\\
\textbf{Hu et al.'s $*$ \cite{hu2019learning}} &81.57$\pm$0.74 &79.35$\pm$2.59 &73.61$\pm$11.86 &	55.84$\pm$11.84 \\
\textbf{Hu et al.'s+Regularization $*$ \cite{hu2019learning}} &82.25$\pm$1.16 &\XSolidBrush&79.53$\pm$1.64 &	66.68$\pm$13.99 \\
\textbf{Liu et al.'s $*$\cite{liu2021improving}}& 82.58$\pm$0.98 &\XSolidBrush& 81.14$\pm$1.25& 80.62$\pm$0.93\\
\textbf{Logit Adjustment $\tau=1.0$ \cite{menon2020long}}& 79.23$\pm$1.68&71.19$\pm$2.09&56.16$\pm$15.45&49.50$\pm$17.11\\
\textbf{LDAM-DRW \cite{}}&78.89$\pm$4.89&66.30$\pm$13.92&65.17$\pm$10.88&53.89$\pm$17.70\\

\textbf{Our method with additional meta set}& \textbf{87.08$\pm$0.09} & \textbf{87.13$\pm$0.04} & \textbf{87.14$\pm$0.08} & \textbf{87.10$\pm$0.05}\\
{\textbf{Our method with dynamic meta set}}& \textbf{87.12$\pm$0.03} & \textbf{87.11$\pm$0.01} & \textbf{87.09$\pm$0.07} & \textbf{87.12$\pm$0.01}\\
\bottomrule
\end{tabular}}\label{sst2_metaset}
\end{table}

{\section{Label-awareness Experiments}\label{sec:label_aware_exp}
In this section, we report label-aware performance of our method in image, text and point cloud classification task. In this setting, \textbf{label-aware} represents that we only use label information in meta dataset to constrain the OT loss and then obtain the updated sample weights. \textbf{our best performance} represents that we use both feature and label information in meta dataset. As summarized in \ref{cifar100_label_aware}, \ref{sst2_label_aware} and \ref{modelnet_label_aware}, all results show that our method could obtain a comparable performance under the \textbf{label-aware} setting, where this label information is free to be obtained and we could reach the goal by only requiring the labels in training dataset obeying a uniform distribution.}

\begin{table}[!ht]
\centering
\caption{{\small{Top-1 errors(\%). Label-aware performance of our method in CIFAR-LT-100 dataset with imbalance factor=100, which is image classification task.}}}
\resizebox{0.4\textwidth}{!}{
\begin{tabular}{c|c}
\toprule[1pt]
\textbf{Method} & \\ \hline
\textbf{Our best performance}& 54.97\\
{\textbf{Label-aware}}& 55.06\\
\bottomrule
\end{tabular}}\label{cifar100_label_aware}
\end{table}

\begin{table}[!ht]
\centering
\caption{{\small{Top-1 accuracy(\%).Label-aware performance of our method in SST-2 dataset under different imbalance factors, which is text classification task.}}}
\resizebox{0.8\textwidth}{!}{
\begin{tabular}{c|cccccc}
\toprule[1pt]
\textbf{Method} & 100 : 1000 & 50 : 1000 & 20 : 1000 & 10 : 1000 \\ \hline
\textbf{Our best performance}& 87.08$\pm$0.09&87.13$\pm$0.04 & 87.14$\pm$0.08 &87.10$\pm$0.05 \\
{\textbf{Label-aware}}& 86.48$\pm$0.29&86.30$\pm$0.37 & 86.19$\pm$0.54 &86.03$\pm$0.46 \\
\bottomrule
\end{tabular}}\label{sst2_label_aware}
\end{table}

\begin{table}[htbp!]
\centering
\caption{{Top-1 instance accuracy \textbf{IA} and class accuracy \textbf{CA}(\%).Label-aware performance of our method in ModelNet 10 dataset under different imbalance factors, which is point cloud classification task.}}
\resizebox{1.0\textwidth}{!}{
\begin{tabular}{c|ccccccccc}
\toprule[1pt]
\multirow{2}{*}{\textbf{Method}} & \multicolumn{2}{c}{1:100} & \multicolumn{2}{c}{4:100} & \multicolumn{2}{c}{8:100} & \multicolumn{2}{c}{10:100} \\
                        & IA          & CA          & IA          & CA          & IA          & CA          & IA           & CA          \\ \hline
\textbf{Our best performance}                     & 83.71$\pm$0.11           & 81.97$\pm$1.59           & 85.15$\pm$1.01           & 83.80$\pm$0.54           & 89.73$\pm$0.32          & 89.09$\pm$0.38          & 90.40$\pm$0.17           & 90.16$\pm$0.56           \\
{\textbf{Label-aware} }                    & 81.36$\pm$0.90          & 80.13$\pm$0.77           & 82.53$\pm$1.00           & 81.61$\pm$1.07           & 86.95$\pm$0.88           & 86.35$\pm$1.02           & 87.33$\pm$0.62          & 86.76$\pm$0.67           \\
\hline
\end{tabular}}\label{modelnet_label_aware}
\end{table}

\end{document}